\documentclass[11pt]{article}
\usepackage[preprint]{acl}

\usepackage{times}
\usepackage{latexsym}
\usepackage[T1]{fontenc}
\usepackage[utf8]{inputenc}
\usepackage{microtype}
\usepackage{inconsolata}
\usepackage{graphicx}

\usepackage{hyperref}       % hyperlinks
\usepackage{url}            % simple URL typesetting
\usepackage{booktabs}       % professional-quality tables
\usepackage{amsfonts}       % blackboard math symbols
\usepackage{nicefrac}       % compact symbols for 1/2, etc.
\usepackage{microtype}      % microtypography
\usepackage{xcolor}         % colors

\usepackage{amsmath}
\usepackage{amsthm,amssymb,multirow}
\usepackage{enumitem}
\usepackage{placeins}
\usepackage{algpseudocode}
\usepackage{algorithm}
\usepackage{listings,tikz}
\usepackage[most]{tcolorbox}
\usepackage{mathbbol}
\tcbuselibrary{minted}
\usepackage{calc}
\usepackage{subcaption}
\usepackage{comment,minted,xcolor,soul}
\sethlcolor{lightblue}
\usepackage{verbatim}
\usepackage{colortbl}
\usepackage{pifont}
\usepackage{fontawesome}

\usepackage{float}
\usepackage{stfloats}

\newcommand{\cmark}{\ding{51}}
\newcommand{\xmark}{\ding{55}}

\newtcblisting{pythoncode}{
  listing engine=minted,
  breakable,
  listing only,
  nobeforeafter,
  minted style=colorful,
  minted language=python,
  enhanced,
  minted options = {
        breaklines=true, 
        fontsize=\small
      },
}

\newtcblisting{javacode}{
  listing engine=minted,
  breakable,
  listing only,
  minted style=colorful,
  minted language=java,
  enhanced,
  minted options = {
        breaklines=true, 
        fontsize=\small
      },
}

\DeclareMathOperator*{\argmax}{arg\,max}

\title{MEC\textsuperscript{3}O: Multi-Expert Consensus for Code Time Complexity Prediction}

\author{
 \textbf{Joonghyuk Hahn\thanks{Equal contribution.}},
 \textbf{Soohan Lim\footnotemark[1]},
 \textbf{Yo-Sub Han}\thanks{Corresponding author.},
 \\
 Department of Computer Science, Yonsei University, Seoul, Republic of Korea,
\\
   \texttt{\{%
   \href{mailto:greghahn@yonsei.ac.kr}{greghahn},%
   \href{mailto:aness1219@yonsei.ac.kr}{aness1219},%
   \href{mailto:emmous@yonsei.ac.kr}{emmous}%
   \}@yonsei.ac.kr}
}

\begin{document}

\maketitle

\begin{abstract}\label{abstract}
Predicting the complexity of source code is essential for software development and algorithm analysis.
Recently, \citet{BaikJHKHK24} introduced CodeComplex for code time complexity prediction.
The paper shows that LLMs without fine-tuning struggle with certain complexity classes.
This suggests that no single LLM excels at every class,
but rather each model shows advantages in certain classes.
We propose \textbf{MEC\textsuperscript{3}O}, a multi-expert consensus system, which extends the multi-agent debate frameworks.
MEC\textsuperscript{3}O assigns LLMs to complexity classes based on their performance and
provides them with class-specialized instructions, turning them into experts.
These experts engage in structured debates,
and their predictions are integrated through a weighted consensus mechanism.
Our expertise assignments to LLMs effectively handle Degeneration-of-Thought,
reducing reliance on a separate judge model,
and preventing convergence to incorrect majority opinions.
Experiments on CodeComplex show that MEC\textsuperscript{3}O
outperforms the open-source baselines, achieving at least 10\% higher accuracy and macro-F1 scores.
It also surpasses GPT-4o-mini in macro-F1 scores on average and demonstrates competitive on-par
F1 scores to GPT-4o and GPT-o4-mini on average.
This demonstrates the effectiveness of
multi-expert debates and a weighted consensus strategy to generate the final predictions.
% Our code is available at
\href{https://github.com/suhanmen/MECO}{\faGithub}
\end{abstract}

\section{Introduction}\label{sec:intro}

\begin{figure}
    \centering
    \includegraphics[width=.99\linewidth]{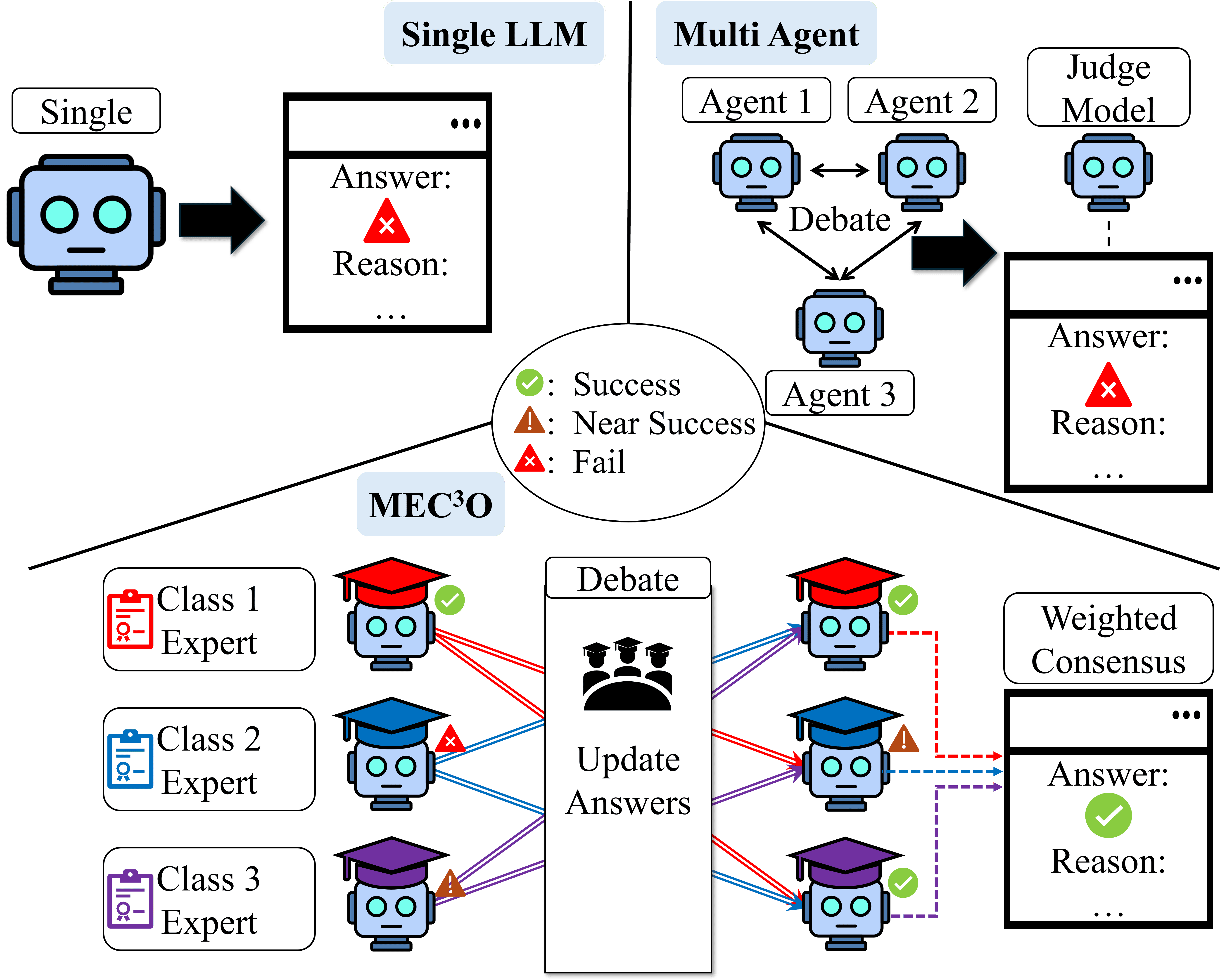}
    \caption{Procedural comparison of Single LLM, multi-agent debate, and MEC\textsuperscript{3}O approaches.}
    \label{fig:intro}
\end{figure}

Time complexity prediction of the source code is a fundamental task in
software development~\cite{SikkaSKUSZ20} and algorithm analysis~\cite{LuGRHSBCDJTLZSZ21,PengZLKHL21,WangZD24},
since it affects the scalability and performance of real-world applications.
Recent advances in large language models~(LLMs) have enabled
automatic code completion, bug fixing, and summarization at near-human levels of fluency~\cite{BouzeniaDP24,HaldarH24,NamMHVM24}.
However, despite the success of single-model paradigms such as GPT-4~\cite{OpenAI}, they
often become locked into a specific line of reasoning, known as
`Degeneration-of-Thought~(DoT)'~\cite{Liang0JW00Y0T24}.
Once a model settles on an initial conclusion,
it may fail to reconsider even when encountering contradictory evidence, limiting overall reliability and coverage.
This procedural distinction among a single LLM, conventional multi-agent debate, and our proposed MEC\textsuperscript{3}O framework is summarized in Figure~\ref{fig:intro}.

Multi-agent debate or self-reflection strategies\citep{Du00TM24,MadaanTGHGW0DPY23, ShinnCGNY23}
attempt to address DoT by involving self-refining or multiple agents that challenge each other's reasoning.
However, these solutions rely on a single underlying LLM architecture or
lack a clear plan for handling conflicting statements for different models.
In some cases, a separate judge model finalizes the outcome, which can overshadow correct but minority opinions.
This problem is observed even in conventional majority votes.
Consequently, improvements are constrained by whichever mechanism to handle the consensus of debate.

Here, we focus on predicting the time complexity of code snippets in multiple programming languages, such as Java and Python.
We define a seven-class categorization scheme:
$O(1)$, $O(\log n)$, $O(n)$, $O(n\log n)$, $O(n^2)$, $O(n^3)$, and $O(2^n)$.
We observe that a single LLM can perform well on one class and poorly on another, suggesting no single model excels across all classes~\cite{BaikJHKHK24}.
We propose \textbf{MEC\textsuperscript{3}O}, a multi-expert consensus system for code time complexity prediction.
Our method builds on the premise that no single LLM excels at every class, but rather each model shows
comparative advantages in certain classes.
MEC\textsuperscript{3}O first uses a small expertise set to identify the LLM that performs the best for each complexity class.
That model then receives a class-specific instruction, becoming an \textit{expert} on that class.
After the debate of these experts, MEC\textsuperscript{3}O utilizes a weighted consensus,
with each expert's vote scaled by its proven reliability for its expertise class.
This setup resolves DoT by letting specialized experts challenge locked-in errors and avoids
relying on a separate judge model.

Empirical results on CodeComplex~\citep{BaikJHKHK24} demonstrate that
MEC\textsuperscript{3}O outperforms all the baselines,
and achieves competitive performance to commercial LLMs such as GPT-4o, GPT-4o-mini, and GPT-o4-mini.
We believe these findings suggest that multi-LLM collaboration,
guided by class-focused experts and weighted consensus,
offers robust performance gains for code complexity prediction and related tasks.

\section{Related Works}\label{sec:related}
\subsection{LLMs for Code Analysis}\label{ssec:rel-LLM-code}
Recently, Large Language Models~(LLMs) have advanced to code-related tasks such as code generation, debugging, and summarization~\cite{AhmedPDB24,CoignionQR24,Zhong0S24}.
They demonstrate remarkable fluency and adaptability, often achieving
strong performance without extensive domain-specific fine-tuning~\cite{AycockB24,LiWLL23, WangWSLCNCZ23, WeiWSBIXCLZ22}.
However, these models can have biases when handling programming styles or domain elements.
In many cases, a single model can be \textit{locked} into its first reasoning path
and fail to revise incorrect inferences, even when new evidence arises~\cite{CreswellSH23, ShinnCGNY23}.
This behavior leads to persistent errors,
which can affect code-related tasks that require deeper reasoning or algorithmic understanding.

\subsection{Code Time Complexity Prediction}\label{ssec:rel-code-tc}
Time complexity prediction has conventionally relied on manual inspection by domain experts~\cite{SikkaSKUSZ20}.
As the number of code snippets scales in size and sophistication,
automated approaches have become increasingly appealing.
There exist several benchmark datasets~\citep{SikkaSKUSZ20,BaikJHKHK24} for this task.
CoRCoD~\cite{SikkaSKUSZ20} is the first benchmark dataset with 932 code snippets covering five complexity classes:
$O(1)$, $O(\log n)$, $O(n)$, $O(n\log n)$, and $O(n^2)$.
Then, extending CoRCoD, \citet{BaikJHKHK24} presented CodeComplex, which consists of 9,800 code snippets of two programming languages, Java and Python,
and covers seven complexity classes:
$O(1)$, $O(\log n)$, $O(n)$, $O(n\log n)$, $O(n^2)$, $O(n^3)$, and $O(2^n)$.
We take CodeComplex to offer a robust testing ground for evaluating how well LLMs generalize across different algorithmic complexities.

\subsection{Multi-Agent Debate}\label{ssec:rel-multi-agent}
Single LLM approaches optimize individual performance~\cite{WeiWSBIXCLZ22,YaoYZS00N23},
whereas multi-agent debate systems leverage diverse LLMs for collaborative reasoning by debate and discussions~\cite{Du00TM24,ChenSB24,Liang0JW00Y0T24,WangWSTS24}.
Basic multi-agent debate frameworks typically involve agents proposing and refuting answers,
with final predictions decided by majority vote~\cite{Du00TM24} or a separate judge~\cite{Liang0JW00Y0T24}.
Other variants introduce confidence-weighted voting~\cite{ChenSB24} or a hierarchical debate across groups~\cite{WangWSTS24}.
However, these methods often suffer from wrong answer propagation,
especially when incorrect majority opinions override correct minority ones.
Judge models also risk bias or failure, as they depend on a single LLM for final decisions.
% Single LLM approaches focus on optimizing its 
% performance~\cite{WeiWSBIXCLZ22,YaoYZS00N23},
% while multi-agent approaches leverage multiple LLMs with diverse reasoning skills~\cite{Du00TM24,ChenSB24,Liang0JW00Y0T24,WangWSTS24}.
% The most basic method concerning multi agents is multi-agent debate~\cite{Du00TM24}.
% Typically, \citet{Liang0JW00Y0T24} uses two agents that iteratively propose and refute answers, with a separate judge determining when a satisfactory solution has been reached. 
% \citet{ChenSB24} suggests a similar approach where each agent produce a confidence score alongside its answer, and then uses confidence-weighted voting to decide the outcome.
% Yet another approach~\citep{WangWSTS24} first groups agents, performs discussions inside each group, and then conducts discussions between groups.
% However, these conventional multi-agent debate approaches all suffer from wrong answer propagation,
% since they cannot fully exploit each agent’s individual strengths and may simply accept incorrect responses.
% It is especially critical for majority voting when the majority generates wrong opinions and the minority generates correct answers.
% A judge model for the final decision still does not fully solve the problem
% as it relies on a single LLM for the judgment.
% This reliance on a single judge model potentially leads to 
% critical errors when the judge’s own performance falters, failing to reflect the consensus of agents' discussions.

Our work addresses these issues by assigning class-specific expert roles to each agent and
using a weighted consensus mechanism that reflects the demonstrated strengths of agents.
This design encourages the refinement of answers when they are contradicted by a domain expert, thereby mitigating DoT and reducing the dilution of correct answers.
In contrast to previous studies, our framework makes consensus decisions without a separate judge,
enhancing robustness and leveraging expert diversity more effectively.

% An agent is more likely to revise its answer if an expert in that class disagrees, which helps avoid locked-in errors.
% This approach not only guards against DoT--since an agent is more likely to revise its response when it is
% contradicted by another agent with demonstrated expertise--but also reduces the dilution of correct answers, thereby mitigating wrong answer propagation.
% Moreover, our weighted consensus combines each agent’s expertise
% with the logit scores of their outputs to make decisions without requiring a separate judge, mitigating over-reliance on the judge model.
% By combining explicit roles with weighted voting,
% we aim to move beyond single LLM performance boundaries and reduce errors of multi-agent debate that arise from unverified or conflicting statements.

\section{Background}\label{sec:prelim-problem}
We briefly overview the time complexity classes
that constitute our prediction targets.
Then, we formalize the overall classification problem, including the dataset,
class labels, and basic notations.
This setup will underpin our multi-agent system described in Section~\ref{sec:method}.

\subsection{Preliminaries}\label{ssec:prelim}
We consider seven time complexity classes as the target labels for code time complexity prediction:
$O(1)$, $O(\log n)$, $O(n)$, $O(n\log n)$, $O(n^2)$, $O(n^3)$, and $O(2^n)$,
where $n$ is the size of the input to the corresponding code snippet.
Below, we briefly describe the classes and explain why we assign expertise at the class level.

\paragraph{Why Class-Level Expertise?\!\!}
Time complexity labels often pose challenges for a single LLM,
which may overfit to superficial loop structures or miss recursion patterns,
leading to persistent errors.
Assigning expertise at the class level and providing class-specific instructions
help mitigate these issues by allowing models to focus on complexity patterns they handle the best.
For instance, an expert specialized in $O(n\log n)$ can accurately
detect sorting algorithms that a general LLM might misinterpret as $O(n)$ or $O(n^2)$.
This targeted expertise reduces errors and improves overall prediction reliability.

\begin{description}[style=unboxed,leftmargin=0cm]
    \item[Constant~($O(1)$):] Code snippets in this category runs in a constant time to their input size.
    These code snippets have constant-time operations or immediate return conditions.
    \item[Logarithmic~$O(\log n)$:] Functions in logarithmic time typically arise in divide-and-conquer scenarios,
    such as binary search.
    They involve recognizing patterns that split the input range in half each step.
    \item[Linear~($O(n)$):] Linear time patterns are the most common in practice,
    where a loop processes each element of the input once.
    \item[Linearithmic~($O(n\log n)$):] Many efficient sorting algorithms and divide-and-conquer
    methods fall into this category.
    \item[Quadratic~($O(n^2)$):] Quadratic runtime typically emerges from a nested loop.
    An expert model for this class can interpret loop bounds and break conditions,
    and then determine the depth of the nested loop to distinguish the complexity class.
    \item[Cubic~($O(n^3)$):] Nested loops of depth three or cubic matrix operations generally indicate $O(n^3)$.
    \item[Exponential~($O(2^n)$):] Exponential code snippets arise in backtracking, exhaustive searches, or subsets generation.
    These show the exponential number of branch factors to the input size.
\end{description}

\begin{figure*}
    \centering
    \includegraphics[width=.98\linewidth]{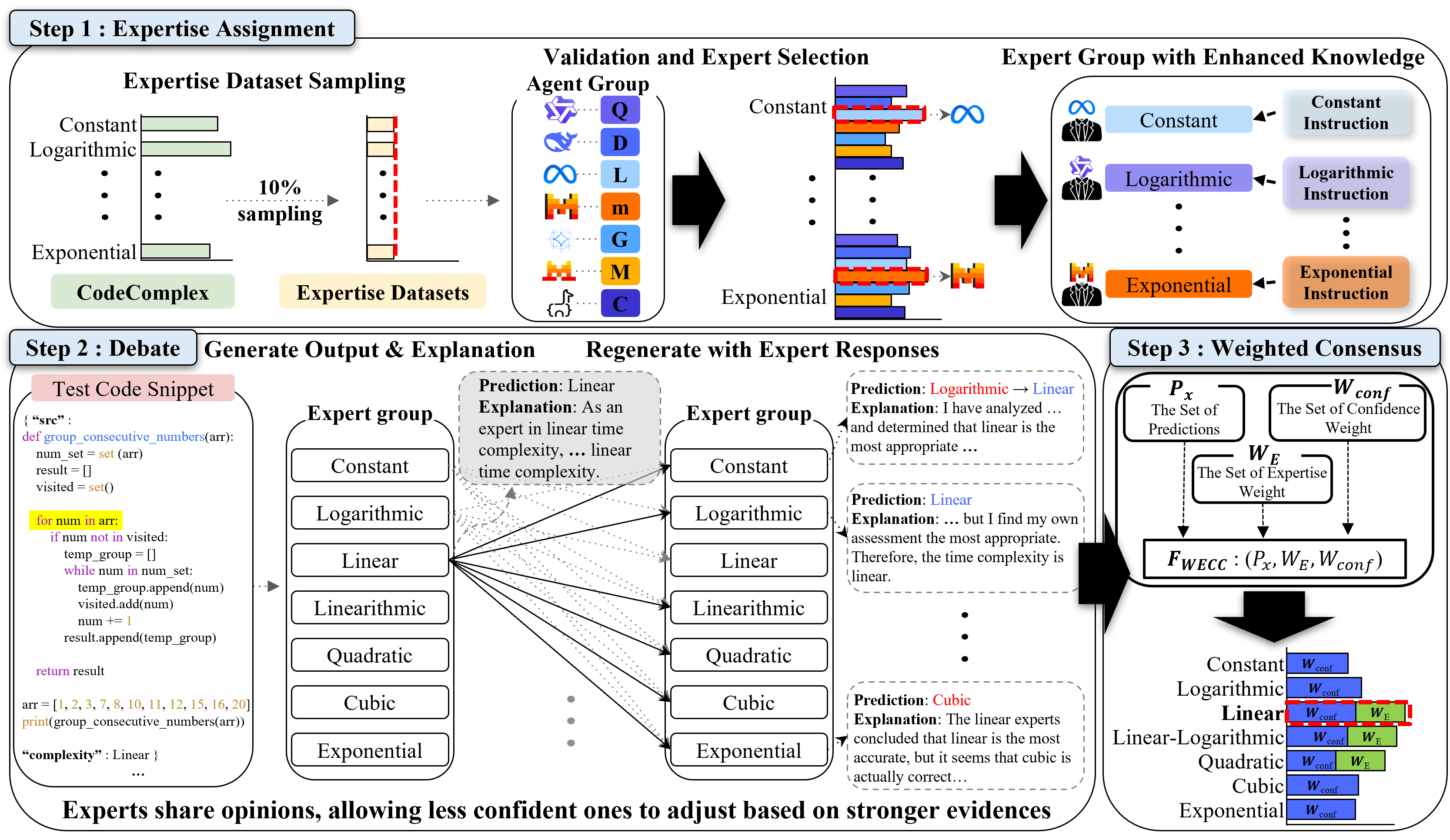}
    \caption{An Overview of MEC\textsuperscript{3}O.
    Step~1: Expertise assignments via model selection and expertise assignments by class-specific instructions.
    Step~2: Multi-expert debates.
    Step~3: Weighted consensus strategy for the final prediction.
    Appendix~\ref{app:debate_procedure} provides a workflow of the debate process.}
    \label{fig:enter-label}
\end{figure*}

\subsection{Problem Setup}\label{ssec:problem}
Let $\mathcal{X}$ be the set of code snippets and $\mathcal{C}$ be the set of time complexity classes defined in Section~\ref{ssec:prelim}.
Our goal is to optimize the reasoning of an LLM~$\mathsf{LLM}$,
which maps each code snippet~$x\in \mathcal{X}$ to a complexity label~$c\in \mathcal{C}$
along with its opinion~$o\in \mathcal{O}$:
\[
\mathsf{LLM}: \mathcal{X} \rightarrow (\mathcal{C},\mathcal{O}).
\]
In our multi-expert framework, we have a set~$M$ of candidate LLMs,
each assigned as an expert for a specific complexity class:
\[
\{M_{c_1}, \ldots, M_{c_7}\}.
\]
For a code snippet~$x$, each expert receives a class-specific instructions~$I_\mathrm{ex}$:
\[
\mathsf{LLM}(x,I_\mathrm{ex}) = (p_i, o_i) \text{ where } p_i\in \mathcal{C} \text{ and } o_i\in \mathcal{O}.
\]
The experts engage in structured debates,
and their outputs are aggregated using a weighted consensus mechanism
to determine the final complexity prediction~$\hat{c}$.
This design addresses the limitation of a single model locking into an incorrect prediction,
ensuring that the final decision is determined by experts in their respective complexity classes.
The full methodology is described in Section~\ref{sec:method}.

\section{Method}\label{sec:method}
\textbf{MEC\textsuperscript{3}O}~(\textbf{M}ulti-\textbf{E}xperts \textbf{C}onsensus for \textbf{C}ode \textbf{Co}mplexity Prediction)
coordinates multiple open-source LLMs into specialized \textit{experts} for different complexity classes,
conducts a \textit{debate} phase to refine opinions, and aggregates final predictions using \textit{weighted consensus}.
Our system draws a clear line between an opinion--the explanatory outputs an experts provides--and
a prediction--the final discrete label an expert offers once debate is complete.
% is a system designed to leverage multiple open-source LLMs for more accurate code time complexity prediction.
% Rather than relying on a single LLM that might be locked into one viewpoint, MEC$^3$O designates \textit{expert} roles
% for each complexity class, coordinates a \textit{debate} among these experts,
% and then produces their final predictions~(or refined opinions) through a \textit{weighted consensus}.
% coordinates a \textit{debate} process, and then produces a final prediction with a \textit{weighted consensus} step.

\subsection{Overall Framework}\label{ssec:method-framework}
Let $\{M_1, M_2, \ldots, M_n\}$ be the set of LLMs, each drawn from distinct model families to enhance diversity.
Given an input code snippet~$x\in \mathcal{X}$,
MEC\textsuperscript{3}O classifies $x$ into $\hat{c}\in \mathcal{C}$ by
proceeding in three steps, 
% 추가
as illustrated in Figure~\ref{fig:enter-label}.
\begin{enumerate}[label=\arabic*)]
\item \textbf{Expertise Assignment.}
For each complexity class, we identify which model excels at with an expertise set,
then assign the best model as the \textit{expert} for that class.
Appendices~\ref{app:Prompt-Constant}--\ref{app:Prompt-Exponential} provide
class-specific instructions for expertise assignments.
\item \textbf{Debate.}
Each designated expert then generates initial complexity predictions for code snippets
using a specialized instruction prompt that highlights its corresponding class.
All experts share their initial opinions.
If an expert notices its overlooked logic in others' outputs,
it may revise its own prediction accordingly.
This exchange supports cross-checking between models and reduces the risk of undetected classification errors.
\item \textbf{Weighted Consensus.}
Once opinions converge, we fuse the final predictions of all experts into a final prediction~$\hat{c}$ via
a weighted consensus function that emphasizes each model's expert class and logit-based confidence.
\end{enumerate}

\subsection{Step~1: Expertise Assignment}\label{ssec:method-expertise}
\paragraph{Expertise Set Partition.}
We randomly sample from CodeComplex
and construct an expertise~(exp) set~$\mathcal{X}_{exp}$,
used exclusively to measure each LLM's performance per class.
For each class~$c$, we treat $c$ as positive and all other classes~$\mathcal{C}\setminus\{c\}$ as negative.
\paragraph{Expert Role Selection.}
We compute a multi-class macro F1 score~\cite{OpitzB19} for every model thereby avoiding trivial strategies such as always predicting the most common class.
Let $\mathrm{F1}_{i,c}$ indicate each LLM~$M_i$'s macro F1 for class~$c$.
We select the LLM with the highest $\mathrm{F1}$ as the expert for $c$:
\[
\mathrm{E}_c = \underset{M_i}{\argmax{}} \mathrm{F1}_{i,c} \text{ for } 1\le i\le n
\]
Ties are permitted but rarely observed.
Each class has a primary expert, ensuring a total $|\mathcal{C}|$ experts.
% Note that we assign an expert to each class, so the number of $\mathrm{Expert}$s is $|\mathcal{C}|$.
% If two or more models share the same top $\mathrm{F1}$, then multiple experts can be assigned to the class.

\subsection{Step~2: Debate}\label{ssec:method-debate}
\paragraph{Single Expert Response.}
We then, proceed to the inference of the test dataset~$\mathcal{X}_{test}$.
For each code snippet~$x\in\mathcal{X}_{test}$\footnote{
We have made clear that $\mathcal{X}_{exp} \cap \mathcal{X}_{test} = \emptyset$.}
, we provide each expert~$M_i$ with a class-specific expertise instruction~$I_{\mathrm{ex},c}$
that informs $M_i$ of its recognized expertise in class~$c$.
Formally, if $\mathrm{E}_c = M_i$, we obtain a prediction-opinion pair:
\[
\{(p_i^x,o_i^x)\}_{i=1}^{|\mathcal{C}|} = M_i(I_{\mathrm{ex},c},x),
\]
where $p_i^x\in \mathcal{C}$ is the expert's initial prediction and
$o_i^x$ is the expert's textual opinion or rationale.
If $p_i^x$ already matches the class~$c$, then by policy,
MEC\textsuperscript{3}O preserves the prediction.
% unless the expert itself chooses to change it after reviewing others in the debate.

\paragraph{Exchange of Opinions.}
Let $\mathcal{P} = \{P_x\mid x \in\mathcal{X}_{test}\}$, where
\[
P_x=\{(p_i^{x},o_i^x)\}_{i=1}^{|\mathcal{C}|}
\]
collects each expert's predictions and opinions for a code snippet~$x$.
We provide each expert with the full set of experts' outputs:
\[
\{((p_i^x)',(r_i^x)')\} = M_i(I_{\mathrm{ex},c}, x \mid \{(p_j^x,\,r_j^x)\}_{j\neq i}^{|\mathcal{C}|}).
\]
The updated prediction~$(p_i^x)'$ may shift to a different class
if the expert discovers hidden loops or recursion from others’ opinions.

\paragraph{Restricted Assents.}
Experts have the option to ignore contradictory suggestions from non-experts if they
strongly trust their own class knowledge.
For instance, if $M_i=\mathrm{E}_{O(1)}$ and $p^x_i= O(1)$,
it may discard feedback from other classes' experts.
This approach prevents the dilution of correct but specialized judgments,
especially when the majority of models disagree due to partial heuristics.
We present the experiments on this in Section~\ref{ssec:ablation} as ablation studies.

\subsection{Step~3: Weighted Consensus}\label{ssec:method-consensus}
After the debate, each expert produces a final prediction~$(p_i^x)'$.
We introduce a Weighted Expertise-Confidence Consensus~(WECC) function:
% which adjusts each expert's class expertise and prediction confidence
% to integrate the predictions into a final decision.
\[
\mathcal{F}_{\text{WECC}}: (P_x, \mathcal{W}_{\text{E}}, \mathcal{W}_{\text{conf}}) \to \hat{c}
\]
where $P_x=\{((p^{x}_i)',(r^x_i)')\}_{i=1}^\mathcal{C}$ is the set of final predictions from all experts,
$\mathcal{W}_{\text{E}} = \{w_{\text{E},i}\}_{i=1}^\mathcal{C}$ represents
the set of expertise weights,
$\mathcal{W}_{\text{conf}} = \{w_{\text{conf},i}\}_{i=1}^\mathcal{C}$ represents
the set of confidence weights,
and $\hat{c}$ is the final prediction.

For a code snippet~$x$, $\mathcal{F}_{\textbf{WECC}}$ operates
by starting to compute the weighted expertise-confidence score~$\mathrm{Score}_x(c)$ for each class~$c$:
\[
\mathrm{Score}_x(c) = \sum_{i=1}^{|\mathcal{C}|} \mathbf{1}((p_i^x)'\equiv c) \cdot w_{i,c}
\]
where the weight~$w_{i,c}$ for each expert~$M_i$ for the class~$c$ is derived as
\[
w_{i,c} = w_{\text{E},i} \cdot w_{\text{conf},i}.
\]
Here, the expertise weight~$w_{\text{E},i}$ prioritizes experts predicting within their assigned class.
Specifically, the expertise weight is defined as
\[
w_{\text{E},i} =
\begin{cases}
    \alpha, & \text{if } (p_i^x)' \equiv c \text{ where $\mathrm{E}_c \equiv M_i$}\\
    \beta,  & \text{otherwise}
\end{cases}
\]
where $(\alpha > \beta)$ ensures that class expertise is preferentially weighted.
This ensures that if $M_i \equiv E_{O(1)}$ and $(p_i^x)' \equiv O(1)$, it receives a strong weight.
$w_{\text{conf},i}$ is derived from the expert's logit confidence score.

The confidence weight~$w_{\text{conf},i}$  reflects the normalized logit score~$l_{i,x}$ or a self-reported confidence probability~$c_{i,x}$ for $(p_i^x)'$.
% where $\tilde{l}_{i,x}$ is the logit score from the expert~$M_i$ for the code snippet~$x$.
Finally, we select
\[
\hat{c} = \argmax_{c\in \mathcal{C}} \mathrm{Score}_x(c).
\]

Because experts who reaffirm their own domain class are granted priority,
MEC\textsuperscript{3}O prevents contradictory or less confident peers from misleading the recognized experts.
This design balances two principles:
each model's superiority in a particular complexity class,
and each model's self-reported or logit-based confidence in the input code snippet.

By preserving expert-aligned predictions, MEC\textsuperscript{3}O offers a robust framework that counters typical single-model pitfalls,
such as ``locking'' onto an initial guess, and typical multi-agent pitfalls, such as
uniform voting that dilutes correct minority views.
The debate and final consensus clearly reflect both class knowledge and snippet-level certainty.
This results in more reliable time complexity classification.

\begin{table*}[hbt]
    \centering
    % \resizebox{0.9\linewidth}{!}{
    \begin{tabular}{lcccccc}
    \toprule
        & \multicolumn{2}{c}{\textbf{Java}} & \multicolumn{2}{c}{\textbf{Python}} & \multicolumn{2}{c}{\textbf{Average}}\\
        \hline
        \textbf{Model}                                     & Acc.           & F1.             & Acc.             & F1.             & Acc.             & F1.   \\
        \midrule
        \textbf{Single LLM}                                &                &                 &                  &                 &                  &       \\
        \hline
        Zero-Shot Instruction~\citep{BrownMRSKDNSSAA20}    & 52.00          & 44.00           & 50.20            & 40.60           & 51.10            & 42.30     \\
        Seven-Shot Instruction~\citep{BrownMRSKDNSSAA20}   & 56.30          & 48.90           & 48.00            & 39.40           & 52.15            & 44.15     \\
        CoT~\citep{WeiWSBIXCLZ22}                          & 54.08          & 45.79           & 52.86            & 44.06           & 53.47            & 44.93     \\
        Self-Consistency~\citep{WangWSLCNCZ23}             & 51.84          & 42.45           & 51.22            & 40.73           & 51.53            & 41.59     \\
        Reflexion~\citep{ShinnCGNY23}                      & 53.47          & 43.89           & 52.24            & 41.96           & 52.86            & 42.93     \\
        \midrule    
        \textbf{Multi-Agent Debate}                        &                &                 &                  &                 &                  &                 \\
        \hline    
        Multiagent~(Majority)~\citep{Du00TM24}             & 54.49          & 50.21          & 52.86             & 49.97           & 53.68            & 50.09           \\
        Multiagent~(Judge)~\citep{Du00TM24}                & 54.90          & 45.10          & 55.30             & 44.60           & 55.10            & 44.85           \\
        MAD~\citep{Liang0JW00Y0T24}                        & 46.33          & 39.72          & 40.00             & 36.36           & 43.17            & 38.04           \\
        RECONCILE~\citep{ChenSB24}                         & 55.92          & 52.79          & 55.31             & 51.11           & 55.62            & 51.95           \\
        CMD~\citep{WangWSTS24}                             & 56.53          & 47.07          & 55.31             & 45.69           & 55.92            & 46.38           \\
        \midrule    
        \textbf{Commercial LLMs}                           &                &                &                  &                  &                  &                 \\
        \hline    
        % GPT-3.5                                   & 54.20          & 52.90          & 45.50             & 44.20           & 49.85            & 48.55           \\
        % GPT-4                                     & 60.00          & 60.60          & 53.80             & 52.70           & 56.90            & 56.65           \\
        % \midrule
        GPT-4o                                             & \textbf{71.72} & 62.22          & 61.09             & 53.08           & \textbf{66.41}   & \textbf{57.65} \\
        GPT-4o-mini                                        & 64.96          & 55.68          & 56.09             & 48.40           & 60.53            & 52.04           \\
        GPT-o4-mini                                        & 65.12          & \textbf{62.31} & \textbf{62.31}    & \textbf{54.23}  & 63.72            & 58.27           \\
        \midrule    
        \textbf{Multi-Expert with Weighted Consensus}      &            &                &                   &                 &                  &                 \\
        \hline
        \addlinespace[2pt]
        \textbf{MEC\textsuperscript{3}O}                   & \textbf{61.02} & \textbf{61.16} & \textbf{57.55}    & \textbf{53.51}  & \textbf{59.29}   & \textbf{57.34}   \\
        \bottomrule
    \end{tabular}
    % }
    \caption{Accuracy~(Acc.) and macro-F1~(F1.) performance of MEC\textsuperscript{3}O and baselines.
    Table~\ref{tbl:app-comparison-weighted} in Appendix~\ref{app:weighted} additionally provides weighted F1 scores.}
    % Baselines include four open-source LLMs and commercial LLMs as single LLM approaches
    % and MAD with a majority vote and a judge model for multi-agent approaches.
    % \textbf{Demon.} denotes zero-shot~(\xmark) and seven-shot~(\cmark) settings.
    \label{tbl:main-meco}
\end{table*}

\subsection{Implementation Details}\label{ssec:method-implementation}
We use representative open-source LLMs both for general-purpose and code-related tasks:
Qwen2.5-Coder-7B-Instruct~(Qwen2.5-Coder)~\citep{HuiYCYLZLZYDYMHRRZL24}, 
deepseek-coder-7b-instruct-v1.5~(Deepseek-Coder)~\citep{GuoZYXDZCBWLLXL24}, 
Ministral-8B-Instruct-2410~(Ministral), 
Meta-Llama-3.1-8B-Instruct~(Llama-3.1)~\cite{Dubeyetal24}, 
Mistral-7B-Instruct-v0.3~(Mistral)~\cite{JiangSMBCCBLLSLLSSLWLS23}, 
codegemma-7b-it~(codegemma)~\cite{Zhaoetal24}, 
CodeLlama-7b-Instruct-hf~(CodeLlama)~\cite{Roziereetal24}.
We designate exactly seven experts in total, as CodeComplex consists of seven time-complexity classes.
Multiple labels may map to the same underlying LLM if that model achieves the top macro F1
on more than one class, but each class-specific expert is prompted with a distinct role instruction.
% For instance, if Qwen2.5-Coder ranks the highest for both $O(\log n)$ and $O(n)$,
% we load Qwen2.5-Coder twice with different expert instructions---one specialized in
% logarithmic class and another focused on linear class.
Appendices~\ref{app:perclass-breakdown} and~\ref{app:expertiese} provide
the performance of all LLMs and the LLM with the highest performance for each class, respectively.

\section{Results and Analysis}\label{sec:result}
\subsection{Experimental Settings}\label{ssec:setting}
We conduct experiments on CodeComplex~\cite{BaikJHKHK24}, which has 4,900 Java and 4,900 Python code snippets,
each annotated with one of seven time-complexity classes.
We follow the official test split in \citet{BaikJHKHK24} to evaluate and then sample 10\% of the data from remaining
to form an expertise dataset for identifying the LLM that excels at each complexity class.
We uniformly sample the same number of samples over all classes.
As discussed in Sections~\ref{ssec:method-expertise} and~\ref{ssec:method-debate},
there is no overlapping data instance from the expertise dataset and test dataset.
All remaining data is not used as our experiments do not conduct fine-tuning.
We use accuracy and macro-F1 for evaluation, reflecting both overall correctness and balance across classes.
We provide baseline details in Appendix~\ref{app:baseline}.
Our experiments are conducted on NVIDIA RTX 3090.

\subsection{Baseline Comparisons}\label{ssec:baseline-comparisons}
Table~\ref{tbl:main-meco} summarizes the performance of MEC\textsuperscript{3}O
under two main categories:
(1) Single LLM approaches and
(2) multi-agent debate methods.
All baselines are evaluated without fine-tuning.
Experiments are also conducted on GPT-4o, GPT-4o-mini, and GPT-o4-mini
to evaluate the competitiveness of MEC\textsuperscript{3}O to powerful commercial LLMs.

Single LLM baselines include zero-, seven-shot instructions\footnote{
We provide the full results of basic zero-, seven-shot instructions in Appendix~\ref{app:single llm performance}.},
Chain-of-Thought~(CoT), Self-Consistency, and Reflexion.
While they enhance reasoning through self-generated explanations or sampling-based agreement, they rely solely on their own outputs.
Consequently, they suffer from DoT and internal bias reinforcement.
MEC\textsuperscript{3}O addresses these limitations by introducing diverse experts
who can challenge, revise, and vote on each other's outputs through structured collaboration.

Multi-agent debate baselines outperform single-LLM methods but still fall short in critical areas,
especially when using the judge model.
Multiagent~(Majority) and Multiagent~(Judge) amplify incorrect majorities
or over-rely on a single judge model, underutilizing class-specific knowledge.
MAD suffers from similar problems when rebuttals are weak and
RECONCILE may discard correct answers by incorrect confidence signals.
CMD struggles when majority opinions across agent groups overpower minority-but-correct views.
MEC\textsuperscript{3}O, in contrast, assigns explicit expertise to agents and use weighted consensus
instead of a judge model to prioritize domain-relevant expertise.
This allows minority--but correct--opinions to dominate when appropriate
and as demonstrated in Table~\ref{tbl:main-meco},
MEC\textsuperscript{3}O achieves about 10\% performance improvements compared to multi-agent debate baselines as well as single LLM approaches
in average both on accuracy and F1 scores.

Notably, MEC\textsuperscript{3}O also demonstrates competitive performance compared to commercial LLMs such as
GPT-4o-mini, GPT-o4-mini, and GPT-4o.
Although, GPT-4o shows better overall performance,
MEC\textsuperscript{3}O achieves better F1 scores than GPT-4o-mini.
This result indicates that when used together appropriately in a multi-agent debate manner,
models with a small parameter size can outperform or be competitive to a single LLM with a much larger parameter size.
We also observe the class-wise performance of MEC\textsuperscript{3}O and baselines in Section~\ref{ssec:error} and Appendix~\ref{app:baseline}.
MEC\textsuperscript{3}O, from Table~\ref{tbl:main-meco}, accomplishes at least 10\% higher accuracy and macro-F1 than any other baselines,
suggesting that class-specific expertise plus a weighted consensus can surpass approaches that rely on a separate judge model. 
Furthermore, Appendix~\ref{app:Computational Cost} provides the computational cost of MEC\textsuperscript{3}O of token generations compared to the baselines.

\begin{table}[hbt]
    \centering
    \resizebox{\columnwidth}{!}{%76
    \begin{tabular}{lccccc}
    \toprule
        & & \multicolumn{2}{c}{\textbf{Java}} & \multicolumn{2}{c}{\textbf{Python}}\\
        \hline
        &                                    \textbf{Judge}                    & Acc.            & F1.            & Acc.            & F1.            \\
        \midrule
        \multirow{2}{*}{Agent-Agent}         & Maj.                            &  54.49          & 50.21          & 52.86           & 49.97          \\
                                             & Weight\textsuperscript{C}       &  52.65          & 49.58          & 55.10           & 51.53          \\
                                             & Weight\textsuperscript{L}       &  53.27          & 50.50          & 55.71           & 52.45          \\
        \multirow{2}{*}{Agent-Expert}        & Maj.                            &  55.10          & 52.13          & 54.49           & 50.01          \\
                                             & Weight\textsuperscript{C}       &  57.76          & 56.03          & 55.31           & 51.28          \\
                                             & Weight\textsuperscript{L}       &  57.14          & 55.40          & 55.51           & 51.56          \\
        \multirow{2}{*}{Expert-Agent}        & Maj.                            &  57.96          & 54.03          & 55.51           & 51.23          \\
                                             & Weight\textsuperscript{C}       &  57.76          & 54.53          & 56.33           & 51.84          \\
                                             & Weight\textsuperscript{L}       &  58.57          & 56.57          & 56.94           & 52.75          \\
        \multirow{2}{*}{Expert-Expert}       & Maj.                            &  57.96          & 54.12          & 56.53           & 52.20          \\
                                             & Weight\textsuperscript{C}       &  58.78          & 57.44          & 56.73           & 53.04          \\
                                             & Weight\textsuperscript{L}       &  \textbf{61.02} & \textbf{61.16} & \textbf{57.55}  & \textbf{53.51} \\
        \bottomrule
    \end{tabular}
    }
    \caption{Ablation studies of MEC\textsuperscript{3}O for expertise assignment and consensus strategy.}
    \label{tbl:ablation-expert-judge}
\end{table}

\subsection{Ablation Studies}\label{ssec:ablation}
Table~\ref{tbl:ablation-expert-judge} illustrates how different configurations in MEC\textsuperscript{3}O--
particularly the timing of expertise assignments and the choice of final decision method--affect predictions.
The comparison for ablation studies involves
whether the model act as \textit{agents} or \textit{experts} before and after the debate phase,
and whether the final prediction is determined through a simple majority vote~(Maj.) or by
weighted consensus function,$\mathcal{F}_{WECC}$.
Weight\textsuperscript{C} and Weight\textsuperscript{L}
denote the performance of $\mathcal{F}_{WECC}$ using
self-reported confidence scores and logit-based scores, respectively.
For instance, the `expert-agent' setting uses experts to generate initial predictions and opinions,
and these experts' outputs are given to the agents.
We apply a simple majority vote or a weighted consensus function on the predictions of agents
to generate the final prediction.
% 추가
We further examine in Table~\ref{tbl:ablation-confidence} the effect of preserving an expert’s 
initial prediction when it aligns with its specialized class.

\begin{table}[hbt]
    \centering
    \resizebox{\columnwidth}{!}{%76
    \begin{tabular}{lccccc}
    \toprule
        \multicolumn{2}{c}{\textbf{MEC\textsuperscript{3}O}}  & \multicolumn{2}{c}{\textbf{Java}} & \multicolumn{2}{c}{\textbf{Python}}\\
        \hline
        \multicolumn{2}{c}{\textbf{Expert Prediction}}    & Acc.   & F1.   & Acc.   & F1.   \\
        \midrule
        \multicolumn{2}{c}{Preserve Prediction}           & \textbf{61.02}  & \textbf{61.16} & \textbf{57.55}  & \textbf{53.51} \\
        \multicolumn{2}{c}{Change Prediction}             & 60.20  & 58.67 & 57.14  & 53.35 \\
        \bottomrule
    \end{tabular}
    }
    \caption{Ablation studies on updates of the expert's prediction that matches its specialized class.}
    \label{tbl:ablation-confidence}
\end{table}

The `Expert-Expert' rows demonstrate that assigning expertise in the initial phase and retaining the expert role when exchanging the answers
consistently yields better performance than other configurations.
A closer look at the remaining rows shows that leveraging experts in generating initial outputs~(Expert-Agent) performs better
than leveraging experts afterward ~(Agent-Expert).
Overall, using experts in both steps improves performance by 8.43\% in average.

MEC\textsuperscript{3}O relies on a weighted consensus rather than a separate judge model.
For ablation studies, we compare $\mathcal{F}_{WECC}$ to conventional majority voting.
Compared to majority voting,
$\mathcal{F}_{WECC}$ grants more influence to opinions of experts,
achieving 3.92\% and 8.71\% improvements in average for accuracy and F1 scores, respectively.
This clearly shows that when specialists are confident, 
they can resolve the problem of incorrect majority opinions,
where minority opinion gives a correct prediction.
The comparison between Weight\textsuperscript{C} and Weight\textsuperscript{L}
suggests that logit-based weights is more appropriate for the weighted consensus,
implying that numeric probabilities from the model's final layer
offer more reliable confidence than the self-generated score in the model's textual output.

\begin{figure*}[ht]
    \centering
    \begin{subfigure}[b]{.32\textwidth}
    \includegraphics[width=\textwidth]{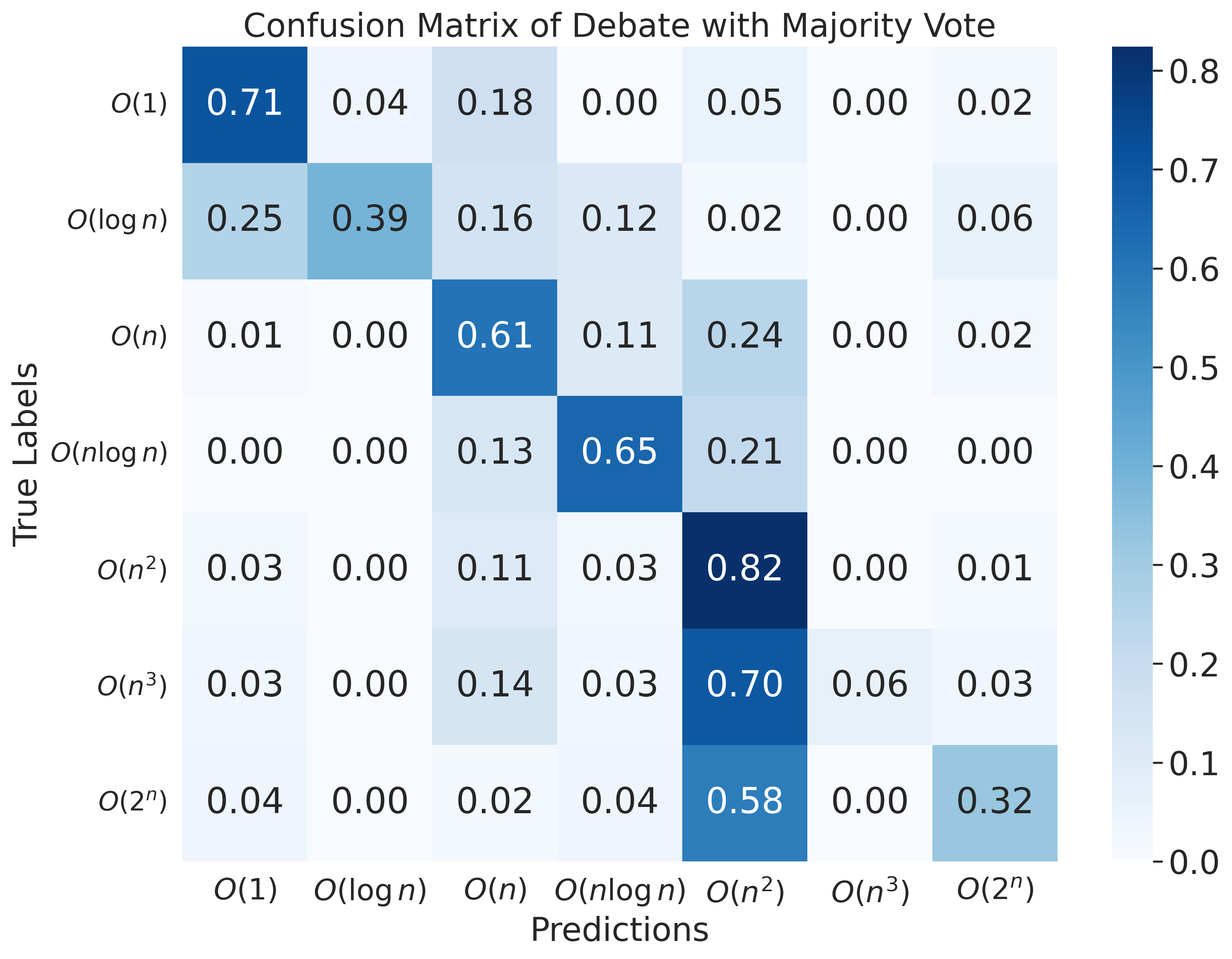}
    \caption{Multiagent~(Majority)}
    \label{fig:java_majority_confusion}
    \end{subfigure}
    \hfill
    \begin{subfigure}[b]{.32\textwidth}
    \includegraphics[width=\textwidth]{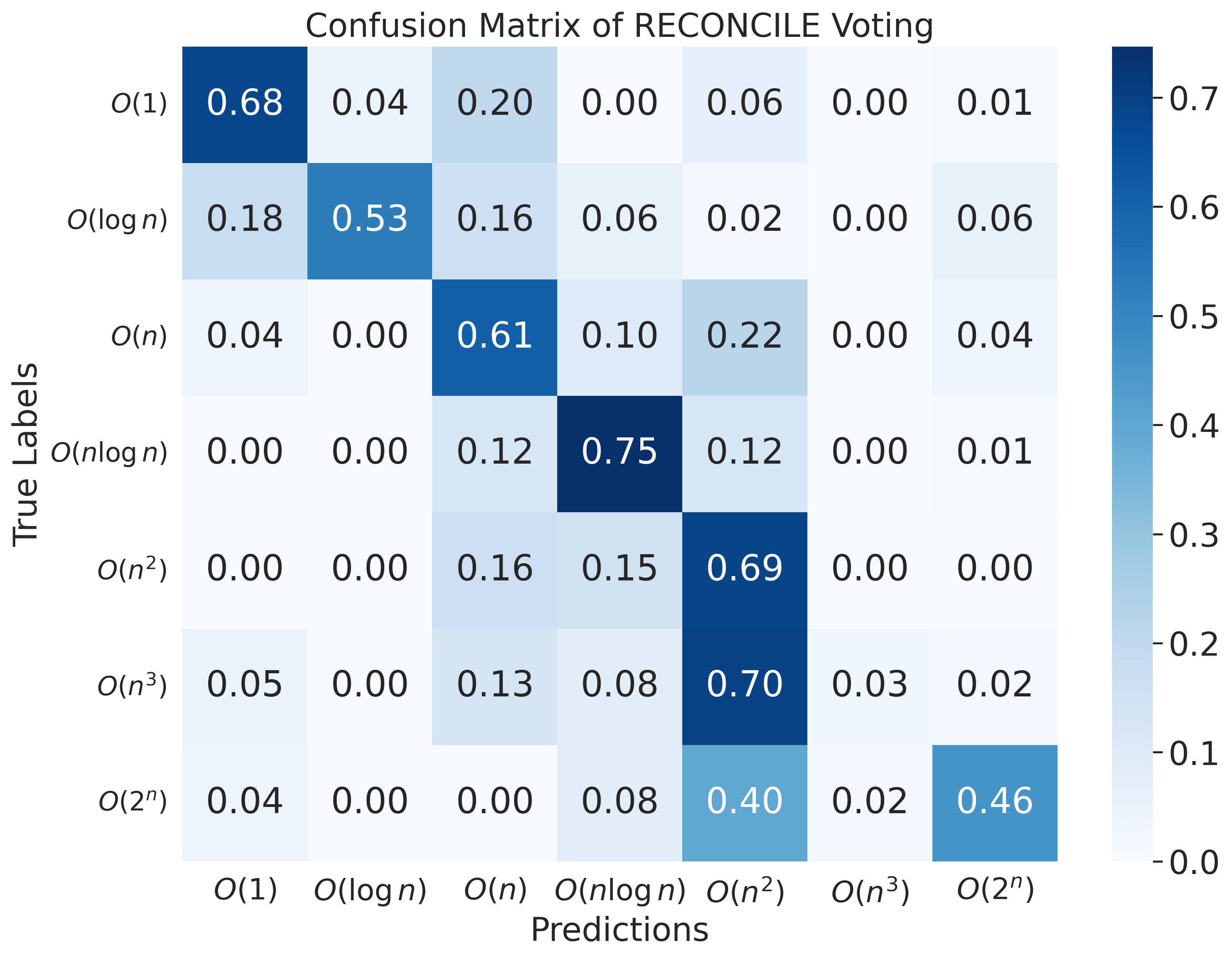}
    \caption{RECONCILE}
    \label{fig:java_reconcile_confusion}
    \end{subfigure}
    \hfill
    \begin{subfigure}[b]{.32\textwidth}
    \includegraphics[width=\textwidth]{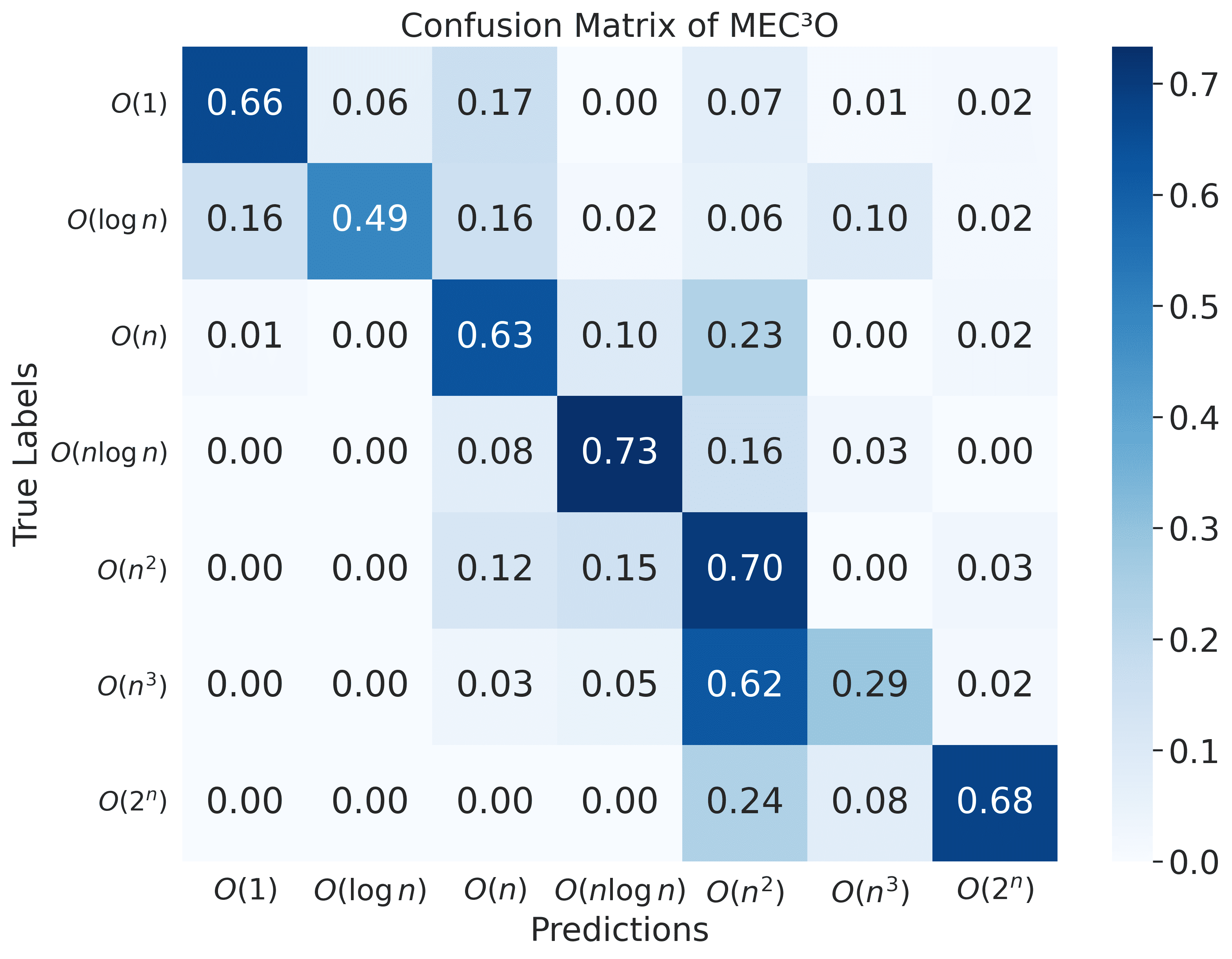}
    \caption{MEC\textsuperscript{3}O}
    \label{fig:java_meco_confusion}
    \end{subfigure}
    \caption{Confusion matrices of Java performance.}
\end{figure*}

Furthermore, we compare the experts who preserve their outputs and those who update their outputs
when their predictions match their expertise class.
Both cases demonstrate similar performance, but preserving the outputs gives a 4.24\% higher F1 score for Java.
This observation implies that experts tend to give more accurate predictions that they are specialized to,
where the opinions of others might provide potential noise.
For instance, when the constant specialist gives outputs of constant predictions,
preserving its outputs give better performance of MEC\textsuperscript{3}O.

\subsection{Error Analysis}\label{ssec:error}
We analyze confusion matrices to assess how MEC\textsuperscript{3}O mitigates class-wise ambiguity and improves
robustness over conventional multi-agent baselines.
For clarity, we focus on two strong baselines---Multiagent~(Majority) and RECONCILE---with full results in Appendix~\ref{app:baseline}.

Multiagent~(Majority), shown in Figure~\ref{fig:java_majority_confusion},
exhibits frequent misclassifications of $O(\log n)$ and $O(n \log n)$ to their adjacent classes
due to subtle structural differences.
MEC\textsuperscript{3}O resolves much of this confusion,
improving $O(\log n)$ and $O(n\log n)$ accuracy by over 10\%.
This reflects the impact of weighted consensus which enforces clearer decision boundaries, especially between neighboring sublinear and linear patterns.
While MEC\textsuperscript{3}O shows on-par performance to RECONCILE in $O(1)$, $O(\log n)$, $O(n)$, $O(n\log n)$, and $O(n^2)$
classes, it achieves significant improvements in $O(n^3)$ and $O(2^n)$ classes in Figures~\ref{fig:java_reconcile_confusion} and~\ref{fig:java_meco_confusion}.
This indicates that
our expert assignment enables better recognition of nesting structures and exponential behaviors, where other systems struggle.

A key strength of MEC\textsuperscript{3}O lies in its smooth and stable performance across all classes.
While the baselines show spikes of strength in some classes and sharp drops in others, MEC\textsuperscript{3}O achieves relatively more stable performance.
Nevertheless, MEC\textsuperscript{3}O does not entirely eliminate all forms of confusion.
Some confusion remains between $O(n^2)$ and $O(n^3)$, which often differ in the depth of nested loops.
LLMs may misinterpret nested loops or their scaling semantics, leading to occasional misclassifications.
Similarly, some ambiguity among $O(1)$, $O(\log n)$, and $O(n)$) classes persists.
Future work may explore more granular expert prompts for close polynomial orders or structured reasoning augmentation to further improve these borderline cases.

\section{Conclusion}\label{sec:conclusion}
We introduce MEC\textsuperscript{3}O, which selects a specialized LLM for each time complexity class using a small expertise dataset,
then after turning the models into \textit{experts} with class-specific instructions,
coordinates these experts through a debate and a weighted consensus.
Our experiments show that this approach mitigates locked-in errors and confusion among classes, achieving at least 10\% higher accuracy and F1 scores,
surpassing all open-source baselines.
MEC\textsuperscript{3}O also proves highly competitive with commercial LLMs, outperforming GPT-4o-mini in F1 score
and achieving competitive performance to GPT-4o and GPT-o4-mini on average.
These results highlight the value of specialized instructions for code time complexity,
the value of multi-expert collaboration, as well as the weighted consensus function that substitutes a judge model.

\section*{Limitations}\label{app:limitation}
\paragraph{Computational Cost.}
MEC\textsuperscript{3}O requires multiple LLMs to process a single input,
making it more expensive than single-LLM approaches.
However, we avoid the additional cost of a separate judge model,
and the system remains feasible with smaller open-source models.
Optimizing expert selection or reducing redundant computations could further improve efficiency.  
\paragraph{Borderline Class Confusion.}
Some complexity classes, such as $O(n^2)$ and $O(n^3)$,
remain difficult to distinguish, even with expert debate and weighted consensus.
MEC\textsuperscript{3}O's performance on $O(n^3)$ is comparable to baselines
while demonstrating notable advances in other classes.
Future improvements could refine how experts interact to handle these edge cases.
\paragraph{Limited Evaluations on Large LLMs.}
MEC\textsuperscript{3}O primarily relies on relatively small open-source LLMs with size 7B--8B parameters.
This choice is reasonable given the high computational cost of using large proprietary models in a multi-agent setup,
but further analysis could investigate whether incorporating larger models as experts would yield additional gains.

%\paragraph{Limitations and Societal Impacts.}
% While MEC\textsuperscript{3}O shows strong performance improvements,
% it requires multiple LLMs per prediction, which increases computational cost compared to single-LLM approaches.
% Some complexity classes, such as $O(n^2)$ and $O(n^3)$, remain difficult to distinguish,
% and our evaluations are limited to models in the 7B–-8B range.
% On the societal side, MEC\textsuperscript{3}O can enhance access to reliable code analysis,
% especially in educational or low-resource settings.
% However, misuse in high-stakes environments without human oversight could lead to erroneous decisions.
% We recommend human-in-the-loop usage and the integration of verification tools to promote responsible deployment.

\bibliography{custom}

\newpage
\onecolumn
\appendix

\section{Expertise Selection}\label{app:expertiese}
We determine model expertise using a small expertise dataset, which is separate from the test set.
Table~\ref{tbl:app-expertise-selection} presents the top-performing LLM for each complexity class across different dataset sampling ratios.  
Qwen2.5-Coder denotes Qwen2.5-Coder-7B-Instruct,
Deepseek-Coder denotes deepseek-coder-7b-instruct-v1.5,
Ministral denotes Ministral-8B-Instruct-2410, and
Llama-3.1 denotes Meta-Llama-3.1-8B-Instruct in Table~\ref{tbl:app-expertise-selection}.
Notably, no single LLM excels in all complexity classes.
Instead, models from different families demonstrate complementary strengths.
For example, Qwen2.5-Coder performs well on logarithmic or linear complexity, whereas Deepseek-Coder and Ministral performs better on linearithmic complexity.
Increasing the expertise dataset size (from 10\% to 30\%) confirms that each model’s expertise remains relatively stable across complexity classes. This consistency allows us to assign models as class-specific experts with high confidence.

\begin{table}[hbt]
\resizebox{\linewidth}{!}{
\begin{tabular}{lcccccccc}
\toprule
Language                & First Ranked Model & constant ($O(1)$)                 & logarithmic ($O(logn)$)     & linear ($O(n)$)             & linearithmic ($O(nlogn)$)   & quadratic ($O(n^2)$) & cubic ($O(n^3)$) & exponential ($O(2^n)$) \\ 
\midrule
\multirow{3}{*}{Python} & 10\%               & Qwen2.5-Coder       & Qwen2.5-Coder & Qwen2.5-Coder & Deepseek-Coder & Qwen2.5-Coder           & Llama-3.1      & Ministral            \\
                        & 20\%               & Qwen2.5-Coder       & Qwen2.5-Coder & Qwen2.5-Coder & Deepseek-Coder & Qwen2.5-Coder           & Llama-3.1      & Qwen2.5-Coder         \\
                        & 30\%               & Qwen2.5-Coder       & Qwen2.5-Coder & Qwen2.5-Coder & Deepseek-Coder & Qwen2.5-Coder           & Llama-3.1      & Ministral            \\\midrule
\multirow{3}{*}{Java}   & 10\%               & Deepseek-Coder & Qwen2.5-Coder & Qwen2.5-Coder & Ministral      & Llama-3.1          & Llama-3.1      & Qwen2.5-Coder             \\
                        & 20\%               & Ministral      & Qwen2.5-Coder & Qwen2.5-Coder & Ministral      & Llama-3.1          & Deepseek-Coder & Qwen2.5-Coder             \\
                        & 30\%               & Deepseek-Coder & Qwen2.5-Coder & Qwen2.5-Coder & Ministral      & Llama-3.1          & Llama-3.1      & Qwen2.5-Coder             \\\bottomrule
\end{tabular}
}
\caption{An overview of the first-ranked model for each time complexity. We exploit the model ranks with varying expertise dataset size~(10\%, 20\%, and 30\%).}
\label{tbl:app-expertise-selection}
\end{table}

\section{Class‐Wise Performance Breakdown for LLMs}\label{app:perclass-breakdown}
We use frequently used open-source LLMs of 7--8B parameter size, as illustrated in Section~\ref{ssec:method-implementation}.
As explained in Section~\ref{ssec:method-implementation},
for each time complexity class,
we select an LLM with the highest performance for the particular class and use the LLM as an expert for that class.
Tables~\ref{tbl:java-classwise} and~\ref{tbl:python-classwise} present the performance of each LLM per class.
Both tables show that Qwen2.5-Coder, Deepseek-Coder, Ministral, and Llama-3.1 performs the best,
each achieving the highest performance for the designated class.
Section~\ref{app:expertiese} presents the model with the top performance per class.

\newpage
\begin{table}[ht]
  \centering
  \renewcommand{\arraystretch}{0.9}
  {\footnotesize        
  \resizebox{0.7\columnwidth}{!}{
    \setlength\tabcolsep{3pt}
    \begin{tabular}{clccccccc}
      \toprule
      Rate & Model & $O(1)$ & $O(\log n)$ & $O(n)$ & $O(n\log n)$ & $O(n^2)$ & $O(n^3)$ & $O(2^n)$ \\
      \midrule
      \addlinespace[0.5em]
      \multicolumn{9}{c}{Java}\\
      \addlinespace[0.1em]
      \midrule
      \midrule
      \multirow{7}{*}{10\%}
        & Qwen2.5-Coder       & 65.67 & \textbf{47.31} & \textbf{62.03} & 62.50 & 50.62 & 10.53 & \textbf{52.43} \\
        & Deepseek-Coder      & \textbf{72.27} & 33.77 & 50.00 & 51.55 & 39.74 & 21.62 & 21.62 \\
        & Llama-3.1           & 46.15 & 23.68 & 41.86 & 46.51 & \textbf{53.62} & \textbf{23.08} & 19.18 \\
        & Ministral           & 69.63 & 27.40 & 53.50 & \textbf{70.37} & 45.45 & 0.00 & 35.29 \\
        & Mistral             & 52.17 & 37.77 & 42.64 & 33.33 & 32.47 & 0.00 & 22.54 \\
        & codegemma           &  9.09 & 14.49 & 15.22 & 11.43 &  2.99 & 0.00 & 0.00 \\
        & CodeLlama           & 31.17 &  3.08 & 22.00 & 21.33 & 17.46 & 0.00 & 0.00 \\
      \midrule
      \multirow{7}{*}{20\%}
        & Qwen2.5-Coder       & 73.61 & \textbf{56.38} & \textbf{53.96} & 51.00 & 38.51 & 2.88 & \textbf{62.56} \\
        & Deepseek-Coder      & 74.56 & 32.89 & 38.91 & 34.94 & 42.23 & \textbf{22.22} & 36.27 \\
        & Llama-3.1           & 56.52 & 34.21 & 37.36 & 27.10 & \textbf{44.44} & 20.65 & 18.18 \\
        & Ministral           & \textbf{74.91} & 25.00 & 47.50 & \textbf{52.36} & 40.82 & 0.00 & 41.92 \\
        & Mistral             & 52.46 & 40.00 & 40.11 & 21.43 & 31.58 &  1.57 & 30.26 \\
        & codegemma           & 13.24 &  9.02 & 22.46 &  5.97 &  1.54 &  0.00 &  0.00 \\
        & CodeLlama           &  9.02 &  7.58 & 21.28 & 13.70 & 18.41 &  0.00 &  4.65 \\
      \midrule
      \multirow{7}{*}{30\%}
        & Qwen2.5-Coder       & 70.77 & \textbf{56.34} & \textbf{54.73} & 55.24 & 44.31 & 16.10 & \textbf{58.50} \\
        & Deepseek-Coder      & \textbf{70.95} & 48.84 & 40.58 & 52.35 & 37.95 & 18.69 & 6.54 \\
        & Llama-3.1           & 53.79 & 27.62 & 36.02 & 38.71 & \textbf{45.80} & \textbf{19.57} & 16.35 \\
        & Ministral           & 70.59 & 27.27 & 45.49 & \textbf{66.88} & 42.58 & 0.00 & 28.94 \\
        & Mistral             & 51.30 & 40.49 & 41.53 & 19.76 & 30.98 &  0.00 & 25.34 \\
        & codegemma           & 10.05 &  5.10 & 16.90 &  7.69 &  3.03 &  1.05 &  0.00 \\
        & CodeLlama           & 15.61 &  7.14 & 21.62 & 22.73 & 19.24 &  2.09 &  7.14 \\
      \bottomrule
    \end{tabular}
  }
  }
  \caption{Per‐class accuracy of LLMs on Java subsets by sampling rate.}
\label{tbl:java-classwise}
\end{table}

\begin{table}[ht]
  \centering
    \renewcommand{\arraystretch}{0.9}
  {\footnotesize       
  \resizebox{0.7\columnwidth}{!}{
    \setlength\tabcolsep{3pt}
    \begin{tabular}{clccccccc}
      \toprule
      Rate & Model & $O(1)$ & $O(\log n)$ & $O(n)$ & $O(n\log n)$ & $O(n^2)$ & $O(n^3)$ & $O(2^n)$ \\
      \midrule
      \addlinespace[0.5em]
      \multicolumn{9}{c}{Python} \\
      \addlinespace[0.1em]
      \midrule
      \midrule
      \multirow{7}{*}{10\%}
        & Qwen2.5-Coder       & \textbf{64.66} & \textbf{57.43} & \textbf{51.81} & 39.66 & \textbf{45.51} & 27.85 & 15.00 \\
        & Deepseek-Coder      & 60.00 & 28.57 & 46.72 & \textbf{51.43} & 42.77 & 38.64 & 17.95 \\
        &      Llama-3.1      & 46.81 & 49.44 & 37.92 & 49.12 & 28.57 & \textbf{40.48} & 14.08 \\
        & Ministral           & 54.29 & 22.54 & 48.35 & 48.84 & 36.29 & 0.00 & \textbf{25.58} \\
        & Mistral             & 47.62 & 51.69 & 39.84 & 5.63  & 23.20 & 9.09 & 0.00  \\
        & codegemma           & 40.96 & 39.75 & 36.36 & 42.55 &  2.74 & 6.15 & 0.00 \\
        & CodeLlama           &  8.70 &  3.03 &  7.89 &  0.00 & 14.74 & 9.09 & 0.00 \\
      \midrule
      \multirow{7}{*}{20\%}
        & Qwen2.5-Coder       & \textbf{64.15} & \textbf{55.56} & \textbf{52.38} & 52.32 & \textbf{42.27} & 35.37 & \textbf{32.37} \\
        & Deepseek-Coder      & 63.07 & 33.12 & 50.68 & \textbf{54.45} & 41.38 & 35.44 & 24.72 \\
        &      Llama-3.1      & 42.78 & 32.68 & 36.84 & 29.38 & 34.55 & \textbf{40.48} & 8.82 \\
        & Ministral           & 60.87 & 22.38 & 47.70 & 32.91 & 40.34 & 0.00 & 28.92 \\
        & Mistral             & 46.33 & 35.84 & 38.39 &  5.59 & 24.69 & 7.63 & 0.00 \\
        & codegemma           & 35.37 & 28.95 & 33.76 & 40.86 &  7.19 & 11.94 & 0.00 \\
        & CodeLlama           & 11.68 & 13.24 & 14.53 & 10.07 & 12.50 & 6.06 & 0.00 \\
      \midrule
      \multirow{7}{*}{30\%}
        & Qwen2.5-Coder       & \textbf{63.75} & \textbf{53.79} & \textbf{49.66} & 52.12 & \textbf{44.86} & 29.75 & 29.23 \\
        & Deepseek-Coder      & 63.25 & 34.04 & 45.48 & \textbf{55.75} & 43.24 & 35.74 & 27.07 \\
        &      Llama-3.1      & 42.14 & 36.21 & 34.11 & 27.24 & 28.39 & \textbf{42.02} & 9.76 \\
        & Ministral           & 61.29 & 26.48 & 44.36 & 31.03 & 38.04 & 0.00 & \textbf{29.92} \\
        & Mistral             & 42.91 & 37.98 & 38.11 &  3.77 & 24.10 &  7.14 &  4.08 \\
        & codegemma           & 36.95 & 26.55 & 34.38 & 41.16 &  4.85 & 12.87 &  0.00 \\
        & CodeLlama           &  8.91 & 10.95 & 13.14 & 10.68 & 11.84 &  8.04 &  0.00 \\
      \bottomrule
    \end{tabular}
  }
  }
  \caption{Per‐class accuracy of LLMs on Python subsets by sampling rate.}
    \label{tbl:python-classwise}
\end{table}

\clearpage

\section{Single LLM Performance}\label{app:single llm performance}
The single LLM category compares four open-source models under zero- and 7-shot conditions.
In the zero-shot setting,
a brief instruction about code time complexity prediction is provided without labeled examples;
in the 7-shot setup, each model sees seven demonstrations, one per class,
along with their assigned time complexities.
A key observation is that single LLMs often lock into incorrect reasoning when encountering less familiar code structures.
%MEC\textsuperscript{3}O addresses this by letting multiple experts challenge such flawed predictions.
\begin{table*}[ht]
    \centering
    \resizebox{0.8\linewidth}{!}{
    % \resizebox{\linewidth}{!}{
    \begin{tabular}{lccccccc}
    \toprule
        & & \multicolumn{2}{c}{\textbf{Java}} & \multicolumn{2}{c}{\textbf{Python}} & \multicolumn{2}{c}{\textbf{Average}}\\
        \hline
        &                              \textbf{Demonstrations}       & Acc.           & F1.             & Acc.             & F1.             & Acc.             & F1.   \\
        \hline
\multirow{2}{*}{Qwen2.5-Coder}         &   \xmark            & 52.00          & 44.00           & 50.20            & 40.60           & 51.10            & 42.30           \\
                                       &   \cmark            & 56.30          & 48.90           & 48.00            & 39.40           & 52.15            & 44.15           \\
\multirow{2}{*}{deepseek-Coder}        &   \xmark            & 36.90          & 33.40           & 43.30            & 39.50           & 40.10            & 36.45           \\
                                       &   \cmark            & 43.90          & 40.70           & 40.60            & 36.70           & 42.25            & 38.70           \\
\multirow{2}{*}{Llama-3.1}             &   \xmark            & 32.20          & 26.70           & 34.70            & 27.50           & 33.45            & 27.10           \\
                                       &   \cmark            & 37.80          & 33.40           & 34.30            & 25.90           & 36.05            & 29.65           \\
\multirow{2}{*}{Ministral}             &   \xmark            & 44.90          & 34.30           & 42.40            & 30.80           & 43.65            & 32.55           \\
                                       &   \cmark            & 46.90          & 37.80           & 43.10            & 35.80           & 45.00            & 36.80           \\
\multirow{2}{*}{Mistral}               &   \xmark            & 33.88          & 26.52           & 31.84            & 21.95           & 32.86            & 24.24           \\
                                       &   \cmark            & 40.00          & 35.29           & 33.67            & 27.24           & 36.84            & 31.27           \\
\multirow{2}{*}{codegemma}             &   \xmark            & 25.42          & 24.02           & 23.67            & 23.10           & 24.55            & 23.56           \\
                                       &   \cmark            & 26.12          & 24.49           & 29.80            & 24.28           & 27.96            & 24.39           \\
\multirow{2}{*}{CodeLlama}             &   \xmark            & 8.78           & 10.22           & 6.94             & 8.25            & 7.86             & 9.24            \\
                                       &   \cmark            & 4.65           & 3.92            & 17.14            & 17.56           & 10.90            & 10.74           \\
    \midrule
    \end{tabular}
    }
    \caption{Accuracy and F1 performance of the baselines. 
    Baselines include seven open-source LLMs.
    % \textbf{Demon.} denotes the usage of demonstrations.
    We use zero-shot~(\xmark) and seven-shot~(\cmark) settings.
    }
    \label{tbl:baseline-comparison}
\end{table*}

\section{Computational Cost}\label{app:Computational Cost}
We measured inference cost in terms of generated tokens. 
As shown in Table \ref{tbl:computational cost}, CoT, Self-Consistency, 
and Reflexion--all single-LLM methods--generate substantially fewer tokens than the multi-agent debate framework. 
Within MAD, two agents engage in a debate, but the judge often makes an early decision to either conclude or continue, which limits the total token count.
MAD generates relatively few tokens compared to other multi-agent approaches, yet it exhibits the lowest performance.
MEC\textsuperscript{3}O, by contrast, generates far fewer tokens than CMD and only marginally more than RECONCILE, yet delivers performance gains of 19.12\% and 9.38 \% 
in average macro-F1 improvements over these methods. 
Thus, MEC\textsuperscript{3}O achieves a clear balance of superior effectiveness and efficiency.

\begin{table}[ht]
    \centering
    \begin{tabular}{cc}
    \hline
    Method                              & Cost~(\# Generated Tokens) \\ \hline
    CoT                                 & 239,452      \\
    Self-Consistency                    & 897,558       \\
    Reflexion                           & 551,253       \\
    RECONCILE                           & 2,012,499      \\
    CMD                                 & 3,219,188      \\
    MAD                                 & 502,751       \\
    MEC\textsuperscript{3}O             & 2,300,898      \\ \hline
    \end{tabular}
    \caption {Total number of tokens processed by each method during inference.}
    \label{tbl:computational cost}
\end{table}

\section{Weighted Scores}\label{app:weighted}
We evaluate MEC\textsuperscript{3}O and baseline models--including
(1) single LLMs and (2) multi-agent approaches--using accuracy and macro F1 in addition to weighted F1.
Unlike macro F1, which treats all classes equally, weighted F1 adjusts for class distributions by assigning different weights based on class frequency.
This provides a more representative evaluation of model performance across the dataset.  

Table~\ref{tbl:app-comparison-weighted} presents the full comparison, including weighted F1.
Notably, the weighted F1-scores are higher than the macro F1-scores.
This is primarily because classes with fewer examples, such as cubic complexity ($O(n^3)$),
contribute less to the overall score compared to more frequent classes.  

% The performance of GPT-3.5 and GPT-4 reports only accuracy and macro F1 as the scores are from \citet{BaikJHKHK24}.
MEC\textsuperscript{3}O achieves the highest weighted F1 across both Java and Python among open-source baselines,
validating its effectiveness in handling diverse complexity classes while maintaining balanced performance.
Furthermore, it demonstrates competitive performance to GPT-4o, GPT-4o-mini, and GPT-o4-mini.

\begin{table*}[ht]
    \centering
    \Large
    \resizebox{\linewidth}{!}{
    \renewcommand{\arraystretch}{1.4}
    \begin{tabular}{lccccccccc}
    \toprule
        & \multicolumn{3}{c}{\textbf{Java}} & \multicolumn{3}{c}{\textbf{Python}} & \multicolumn{3}{c}{\textbf{Average}}\\\hline
                                                                          & Acc.           & F1.                & Weighted F1.       & Acc.              & F1.             & Weighted F1.               & Acc.             & F1.                & Weighted F1      \\\midrule
\multicolumn{10}{l}{\textbf{Single LLM}} \\\hline
zero-shot Instruction~\citep{BrownMRSKDNSSAA20}                           & 52.00          & 44.00              & 51.80              & 50.20             & 40.60           & 49.70                      & 51.10            & 42.30              & 50.75                   \\
seven-shot Instruction~\citep{BrownMRSKDNSSAA20}                          & 56.30          & 48.90              & 56.60              & 48.00             & 39.40           & 47.20                      & 52.15            & 44.15              & 51.90                   \\
CoT~\citep{WeiWSBIXCLZ22}                                                 & 54.08          & 45.79              & 53.62              & 52.86             & 44.06           & 52.99                      & 53.47            & 44.93              & 53.31               \\
Self-Consistency~\citep{WangWSLCNCZ23}                                    & 51.84          & 42.45              & 50.28              & 51.22             & 40.73           & 49.69                      & 51.53            & 41.59              & 49.99               \\
Reflexion~\citep{ShinnCGNY23}                                             & 53.47          & 43.89              & 51.70              & 52.24             & 41.96           & 51.38                      & 52.86            & 42.93              & 51.54               \\
                                                            
                                                            \midrule    
\multicolumn{10}{l}{\textbf{Multi-Agent Debate}} \\\hline 
            Multiagent~(Majority)~\citep{Du00TM24}                        & 54.49          & 50.21              & 52.42              & 52.86             & 49.97           & 51.76                      & 53.68            & 50.09              & 52.09              \\
            Multiagent~(Judge)~\citep{Du00TM24}                           & 54.90          & 45.10              & 53.70              & 55.30             & 44.60           & 54.40                      & 55.10            & 44.85              & 54.05              \\
            MAD~\citep{Liang0JW00Y0T24}                                   & 46.33          & 39.72              & 46.51              & 40.00             & 36.36           & 43.77                      & 43.17            & 38.04              & 45.14              \\
            RECONCILE~\citep{ChenSB24}                                    & 55.92          & 52.79              & 53.95              & 55.31             & 51.11           & 54.33                      & 55.62            & 51.95              & 54.14              \\
            CMD~\citep{WangWSTS24}                                        & 56.53          & 47.07              & 55.61              & 55.31             & 45.69           & 55.02                      & 55.92            & 46.38              & 55.32              \\
            
            \midrule    
\multicolumn{10}{l}{\textbf{Commercial LLMs}} \\\hline
        GPT-4o                                    & \textbf{71.72} & 62.22          & --                 & 61.09             & 53.08           & --                 & \textbf{66.41}   & \textbf{57.65} & --                 \\
        GPT-4o-mini                               & 64.96          & 55.68          & --                 & 56.09             & 48.40           & --                 & 60.53            & 52.04          & --                 \\
        GPT-o4-mini                               & 65.12          & \textbf{62.31} & --                 & \textbf{62.31}    & \textbf{54.23}  & --                 & 63.72            & 58.27          & --                 \\
        \midrule
\multicolumn{10}{l}{\textbf{Multi-Expert with Weighted Consensus}} \\\hline
        \textbf{MEC\textsuperscript{3}O}                                 & \textbf{61.02} & \textbf{61.16}     & \textbf{61.65}     & \textbf{57.55}    & \textbf{53.51}  &  \textbf{56.30}            & \textbf{59.29}   & \textbf{57.34}     & \textbf{58.98}      \\
        \bottomrule
    \end{tabular}
    }
    \caption{Accuracy, F1, and weighted F1 performance of MEC\textsuperscript{3}O and baselines.
    }
    % \textbf{Demon.} denotes zero-shot~(\xmark) and seven-shot~(\cmark) settings.}
    \label{tbl:app-comparison-weighted}
\end{table*}

\section{Statistical Significance of MEC\textsuperscript{3}O}\label{app:statistical_significance}
We evaluate whether MEC\textsuperscript{3}O's performance improvements over baselines in Table~\ref{tbl:app-comparison-weighted}
are statistically significant by using a \textit{paired sign test}~\citep{Demsar06}, a non-parametric test suitable for single-run evaluations.
We compare MEC\textsuperscript{3}O to each baseline across six evaluation settings: Accuracy, macro-F1, and weighted-F1 for both Java and Python.

For each baseline, we count the number of settings where MEC\textsuperscript{3}O outperforms it.
Under the null hypothesis that both methods are equally good, the probability of a win on any task is 0.5.
The one-sided $p$-value is calculated as:
\[
p = \sum_{i=k}^{n} \binom{n}{i} \cdot 0.5^n
\]
where $n = 6$ and $k$ is the number of wins. 
As shown in Table~\ref{tbl:significance}, 
MEC\textsuperscript{3}O achieves perfect wins against every baseline across all six settings.

\begin{table}[ht]
\centering
\small
\begin{tabular}{lccc}
\toprule
\textbf{Baseline} & \textbf{Wins ($k$)} & \textbf{Losses} & \textbf{$p$-value} \\
\midrule
Zero-Shot Instruction     & 6 & 0 & 0.016* \\
Seven-Shot Instruction    & 6 & 0 & 0.016* \\
CoT                       & 6 & 0 & 0.016* \\
Self-Consistency          & 6 & 0 & 0.016* \\
Reflexion                 & 6 & 0 & 0.016* \\
Multiagent (Majority)     & 6 & 0 & 0.016* \\
Multiagent (Judge)        & 6 & 0 & 0.016* \\
MAD                       & 6 & 0 & 0.016* \\
RECONCILE                 & 6 & 0 & 0.016* \\
CMD                       & 6 & 0 & 0.016* \\
% GPT-3.5                   & 6 & 0 & 0.016* \\
% GPT-4                     & 4 & 2 & 0.219 \\
\bottomrule
\end{tabular}
\caption{Paired sign test comparing MEC\textsuperscript{3}O to each baseline across six task-metric pairs.
* denotes statistical significance at $\alpha = 0.05$.}
\label{tbl:significance}
\end{table}

MEC\textsuperscript{3}O consistently outperforms all baselines across all six metrics,
achieving perfect wins in every comparison~($p = 0.016$).
This significantly rejects the null hypothesis at the 5\% level.
These results support the claim that MEC\textsuperscript{3}O offers consistent and statistically meaningful gains in accuracy, macro, and weighted F1 scores across languages.

\section{Baselines}\label{app:baseline}
\begin{enumerate}[label=\arabic*)]
\item \textbf{Chain of Thought~(CoT)~\cite{WeiWSBIXCLZ22}:}
A brief instruction about code time complexity prediction is provided without labeled examples, and the model is encouraged to generate intermediate reasoning steps to reach a final answer.
\item \textbf{Self-Consistency~\cite{WangWSLCNCZ23}:}
It extends CoT by sampling multiple reasoning paths and selecting the most consistent answer through majority voting.
\item \textbf{Reflexion~\cite{ShinnCGNY23}:}
Reflexion enables agents to improve through verbal self-reflection rather than weight updates. After each trial, agents generate natural language feedback based on outcomes and store it in memory to guide future decisions.
\item \textbf{Multiagent~\cite{Du00TM24}:}
Each agent first generates its own response independently. Then, during the debate phase, agents exchange and reason over the responses of others, excluding their own. The final answer is determined via a majority vote or a separate judge model.
\item \textbf{MAD~\cite{Liang0JW00Y0T24}:}
The proposed framework consists of two agents--an affirmative and a negative one--who engage in a debate. The affirmative agent first generates a response, after which the negative agent produces a counter-argument. A separate judge model then decides whether to proceed with additional debate rounds or to conclude with a final decision based on the current arguments.
\item \textbf{RECONCILE~\cite{ChenSB24}:}
Each agent generates an answer with a confidence score. After several rounds of discussion, a final answer is selected via confidence-weighted voting, where each agent's vote is weighted by its calibrated confidence.
\item \textbf{CMD~\cite{WangWSTS24}:}
Agents are divided into two groups. While agents within the same group share both answers and explanations, only answers are exchanged across groups. A majority vote determines the final answer, with a secretary agent resolving ties if necessary.
\end{enumerate}

\subsection{CoT Analysis}\label{app:CoT Analysis}
CoT prompting guides the model to lay out its intermediate reasoning steps before arriving at a final answer.
By making its ``thinking'' explicit, CoT can uncover hidden reasoning patterns that lead to correct solutions.
However, CoT is susceptible to DoT.
Once the model commits to an initial reasoning trajectory,
it rarely backtracks to correct flawed premises, causing early mistakes to propagate all the way through the final judgment.

\begin{figure*}[ht]
    \centering
    % \begin{subfigure}[b]{.49\textwidth}
    \begin{subfigure}[b]{.45\textwidth}
    \includegraphics[width=\textwidth]{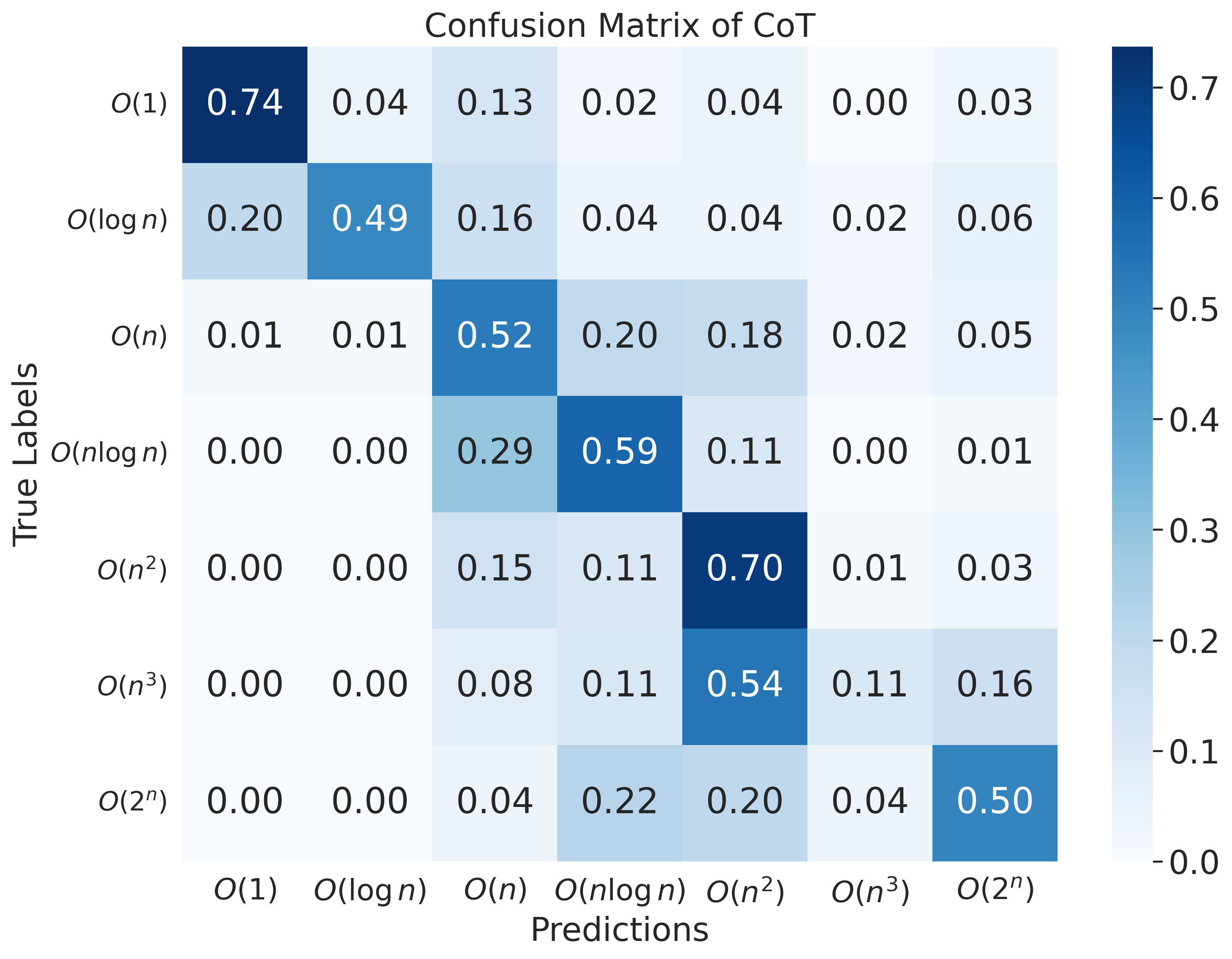}
    \caption{Java}
    \label{fig:CoT-java}
    \end{subfigure}
    \hfill
    % \begin{subfigure}[b]{.49\textwidth}
    \begin{subfigure}[b]{.45\textwidth}
    \includegraphics[width=\textwidth]{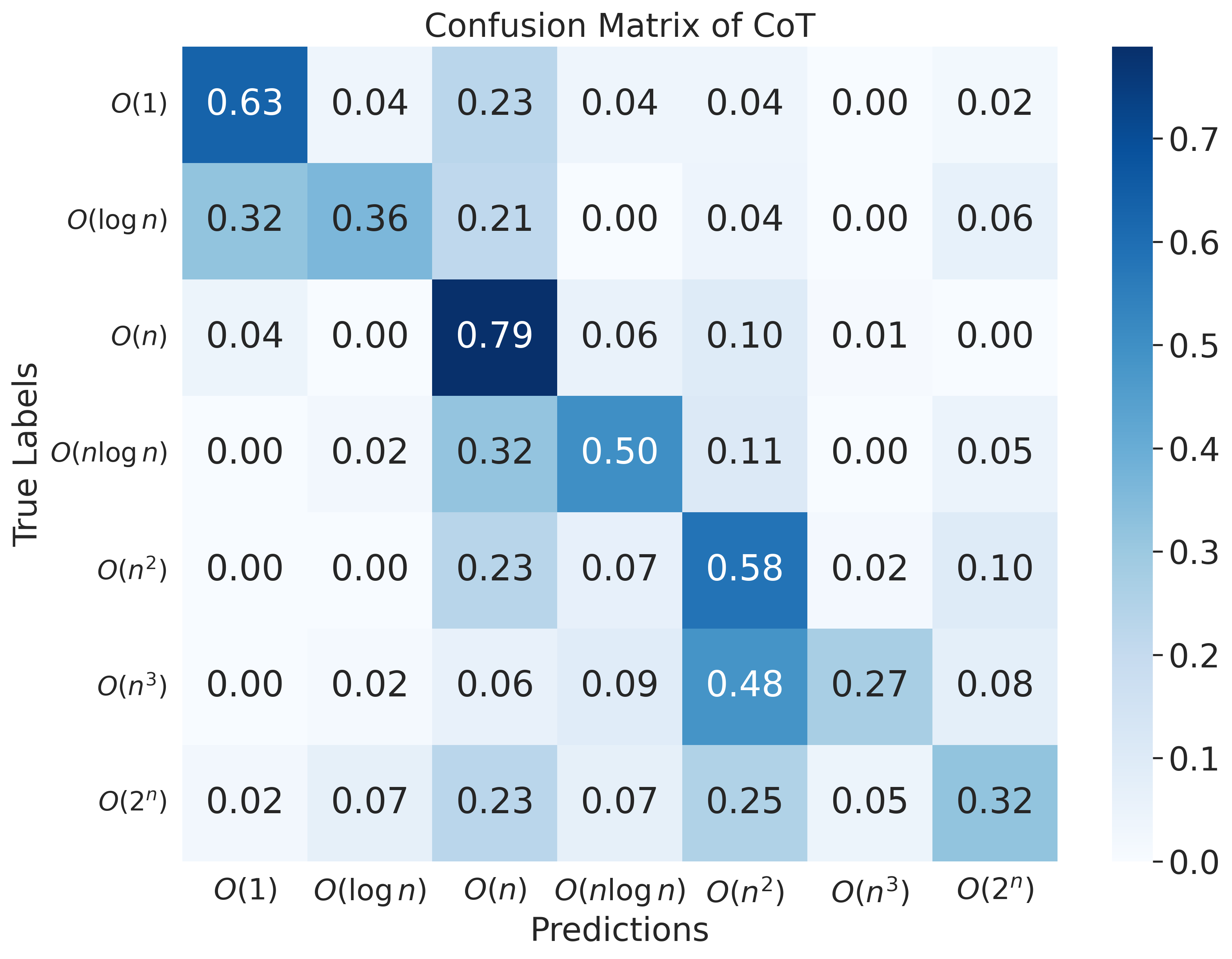}
    \caption{Python}
    \label{fig:CoT-python}
    \end{subfigure}
    \caption{CoT confusion matrices for Java and Python.}
    \label{fig:CoT-analysis}
    \hfill
\end{figure*}

The confusion matrices in Figure~\ref{fig:CoT-analysis} provide further insight into this behavior.
For both Java and Python, CoT demonstrates strong accuracy on simpler classes such as $O(1)$ and $O(n)$,
and particularly excels on $O(n^2)$ in the Java setting with an accuracy of 70\%.
However, the model consistently struggles with higher complexity classes like $O(n^3)$ and $O(2^n)$,
which are frequently misclassified into adjacent lower-complexity classes--for instance, $O(n^3)$ is often predicted as $O(n^2)$.
These errors reveal a tendency to default to more familiar or simpler reasoning patterns when the structural signals are ambiguous or sparse.
This supports the interpretation that CoT's explicit reasoning process, while beneficial for transparency,
can also lead to rigid thinking paths that reinforce early misconceptions rather than revise them--underscoring its vulnerability to DoT.

\subsection{Self-Consistency Analysis}\label{app:Self-Consistency Analysis}
Self-Consistency reduces output variability by sampling several independent reasoning paths and choosing the most frequent answer.
This voting approach curbs random fluctuations, but reinforces the model’s built-in reasoning biases. As a result, performance gains tend to be confined to problem types that already match those biases, rather than fostering truly robust reasoning across diverse cases.

\begin{figure*}[ht]
    \centering
    % \begin{subfigure}[b]{.49\textwidth}
    \begin{subfigure}[b]{.45\textwidth}
    \includegraphics[width=\textwidth]{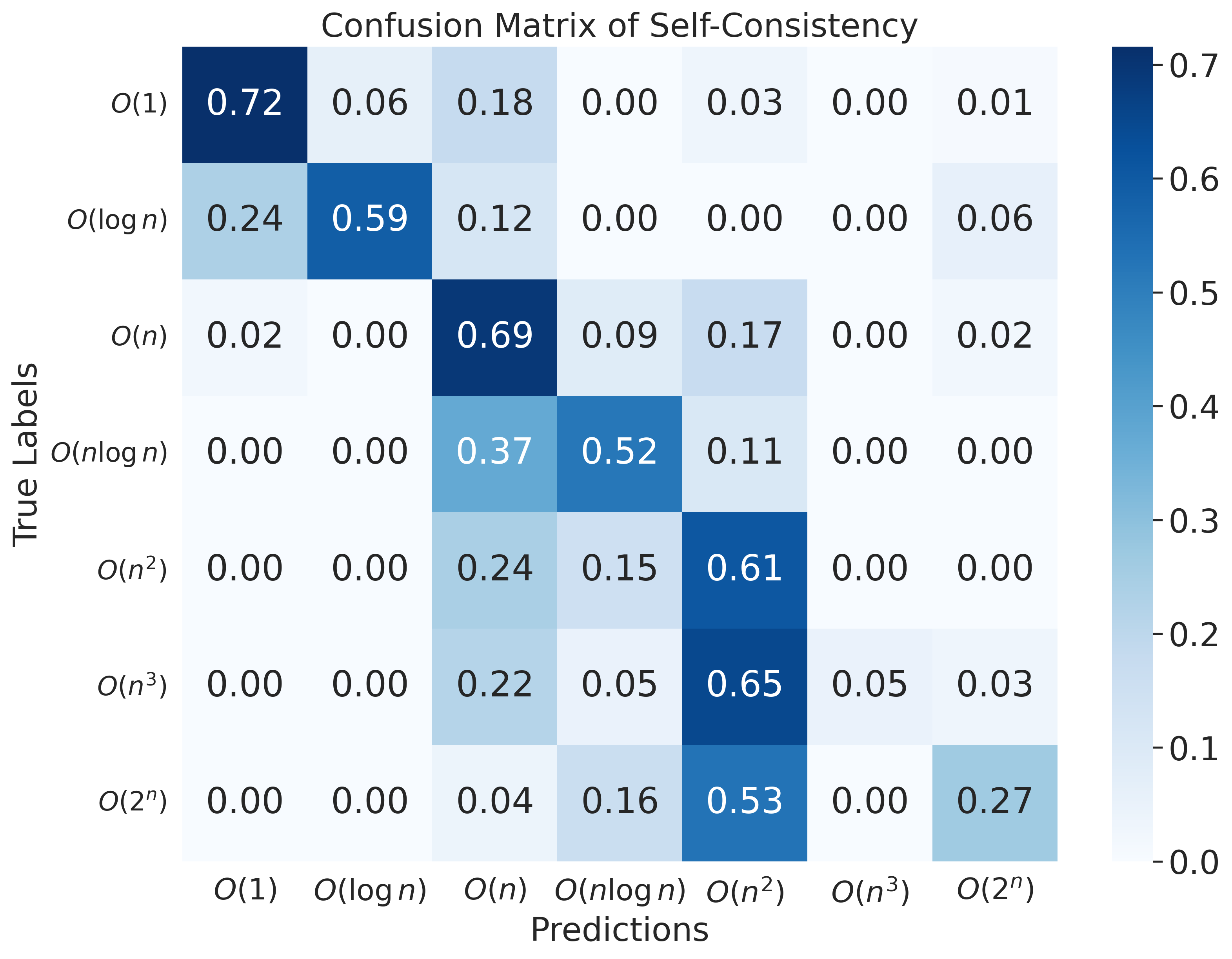}
    \caption{Java}
    \label{fig:SC-java}
    \end{subfigure}
    \hfill
    % \begin{subfigure}[b]{.49\textwidth}
    \begin{subfigure}[b]{.45\textwidth}
    \includegraphics[width=\textwidth]{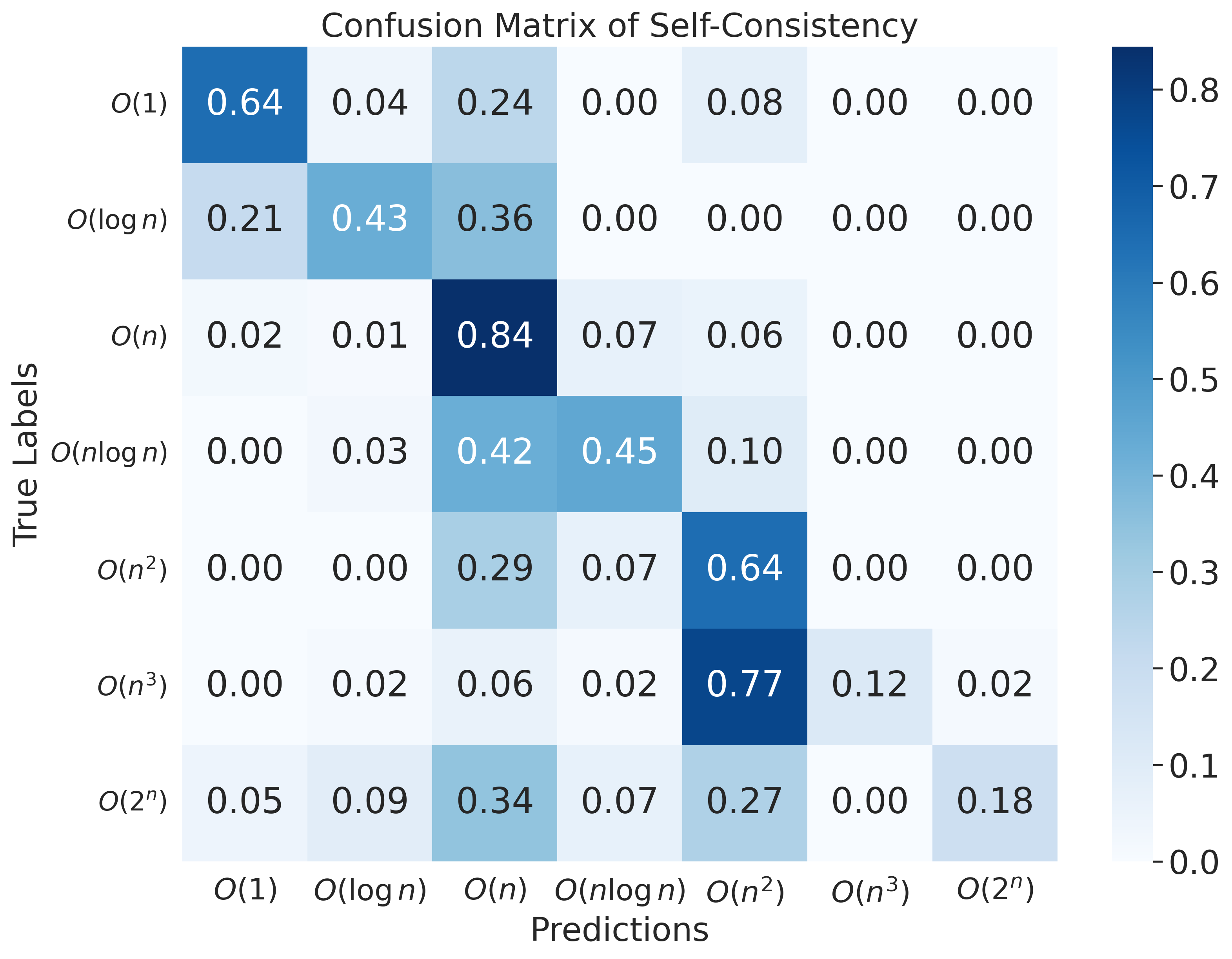}
    \caption{Python}
    \label{fig:SC-python}
    \end{subfigure}
    \caption{Self-Consistency confusion matrices for Java and Python.}
    \label{fig:SC ALL}
    \hfill
\end{figure*}

The confusion matrices in Figure~\ref{fig:SC ALL} provide further evidence of this bias reinforcement. While Self-Consistency improves performance for frequently occurring and structurally simple classes such as $O(1)$ and $O(n)$, it performs notably worse on higher complexity classes like $O(n^3)$ and $O(2^n)$. In both Java and Python, these classes are frequently misclassified into simpler categories, indicating that the majority-vote mechanism amplifies the model’s tendency to converge on safe, low-complexity predictions. As a result, Self-Consistency may stabilize outputs but fails to recover from biased reasoning paths, particularly when confronting edge-case or rare complexity patterns.

\subsection{Reflexion Analysis}\label{app:Reflexion Analysis}
Reflexion extends CoT paradigm by introducing a self-reflection loop. 
After generating an initial inference, the model produces 
a natural language summary of its own performance and stores it in memory. 
These stored reflections serve as semantic signals to guide subsequent attempts. 
However, the loop can become fixed to the initial inference, 
meaning that early errors persist.
When feedback originates from similarly biased reasoning or lacks sufficient corrective cues, the model fails to overcome those errors, leading to a failure mode known as DoT.

\begin{figure*}[ht]
    \centering
    % \begin{subfigure}[b]{.49\textwidth}
    \begin{subfigure}[b]{.45\textwidth}
    \includegraphics[width=\textwidth]{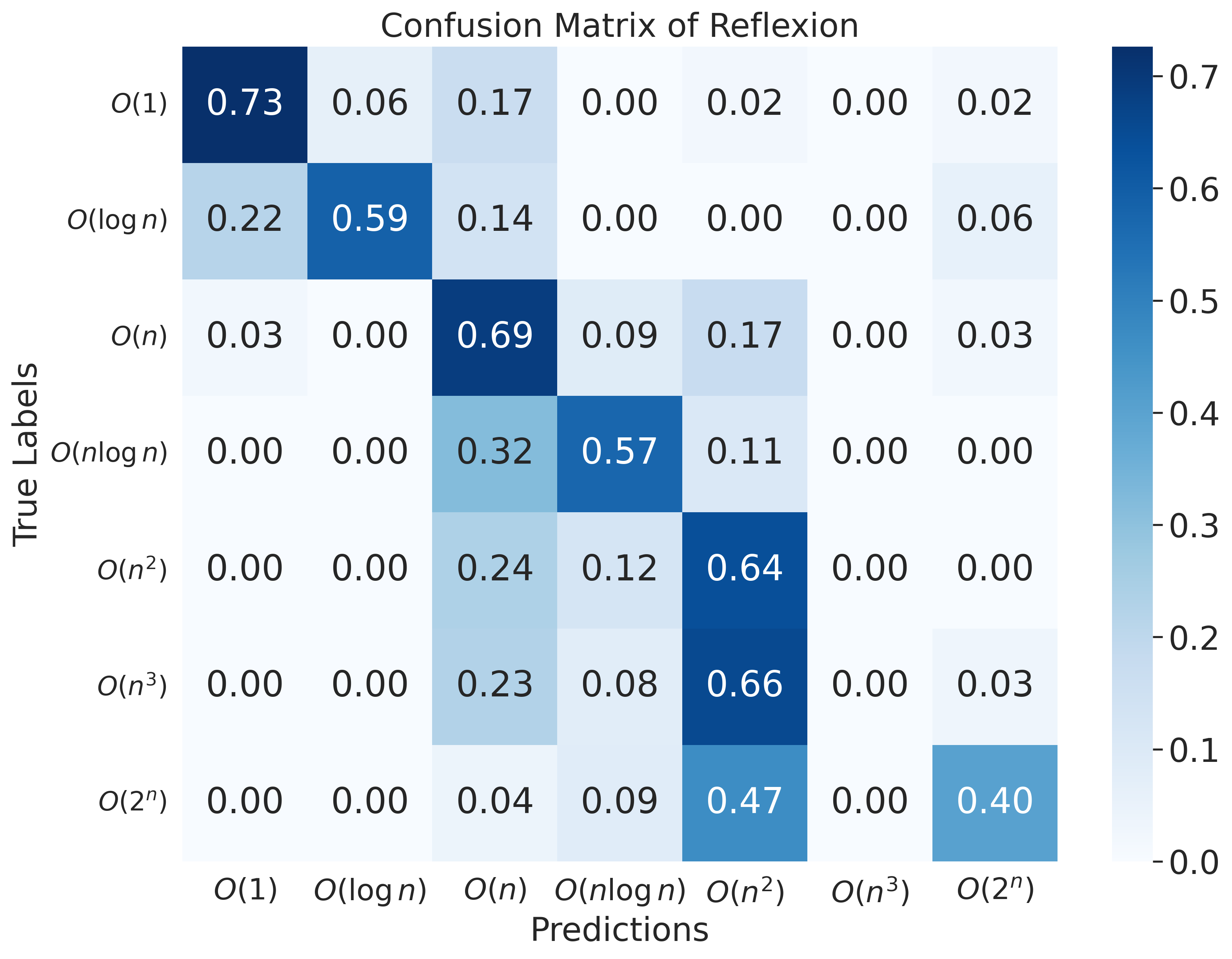}
    \caption{Java}
    \label{fig:Reflexion-java}
    \end{subfigure}
    \hfill
    % \begin{subfigure}[b]{.49\textwidth}
    \begin{subfigure}[b]{.45\textwidth}
    \includegraphics[width=\textwidth]{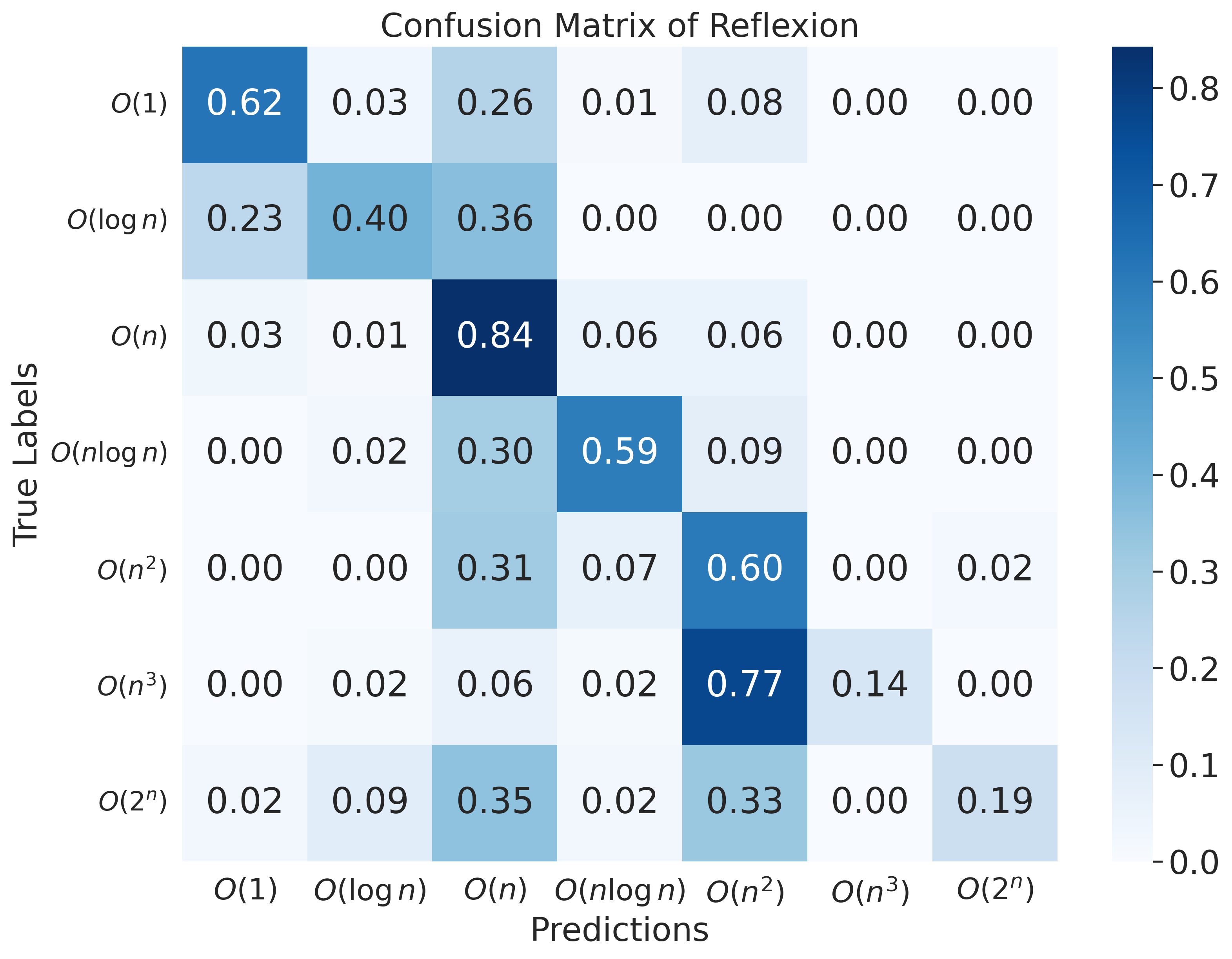}
    \caption{Python}
    \label{fig:Reflexion-python}
    \end{subfigure}
    \caption{Reflexion confusion matrices for Java and Python.}
    \label{fig:Reflexion-ALL}
    \hfill
\end{figure*}

Figure~\ref{fig:Reflexion-ALL} shows that Reflexion offers marginal improvements in certain complexity classes, such as $O(n^2)$ and $O(n^3)$, particularly in the Python setting. However, its performance remains inconsistent across the full range of complexity levels. For instance, in the Java confusion matrix, Reflexion achieves moderate accuracy on $O(2^n)$, but still exhibits substantial confusion with neighboring polynomial classes. Notably, predictions for $O(n^3)$ and $O(2^n)$ often spill into $O(n^2)$, indicating that the model's reflective revisions do not reliably correct initial biases. Instead, reflection tends to reinforce early reasoning paths unless strong corrective signals are available. This observation supports the concern that, despite offering semantically richer feedback than CoT, Reflexion may still perpetuate flawed inferences when the internal reflections are themselves biased or uninformative.

\subsection{Multiagent Analysis}\label{app:Multiagent analysis}
Multiagent~(Majority) relies on conventional majority voting to finalize predictions. While this approach enables simple collaborative reasoning, it does not leverage the specific expertise of each model's class. Consequently, incorrect majority opinions get amplified, and correct answers proposed by minority agents are often ignored, limiting overall reliability.

\begin{figure*}[ht]
    \centering
    % \begin{subfigure}[b]{.49\textwidth}
    \begin{subfigure}[b]{.45\textwidth}
    \includegraphics[width=\textwidth]{Figures/java_multiagent_majority_confusion.png}
    \caption{Java}
    \label{fig:Multiagent~(Majority)-java}
    \end{subfigure}
    \hfill
    % \begin{subfigure}[b]{.49\textwidth}
    \begin{subfigure}[b]{.45\textwidth}
    \includegraphics[width=\textwidth]{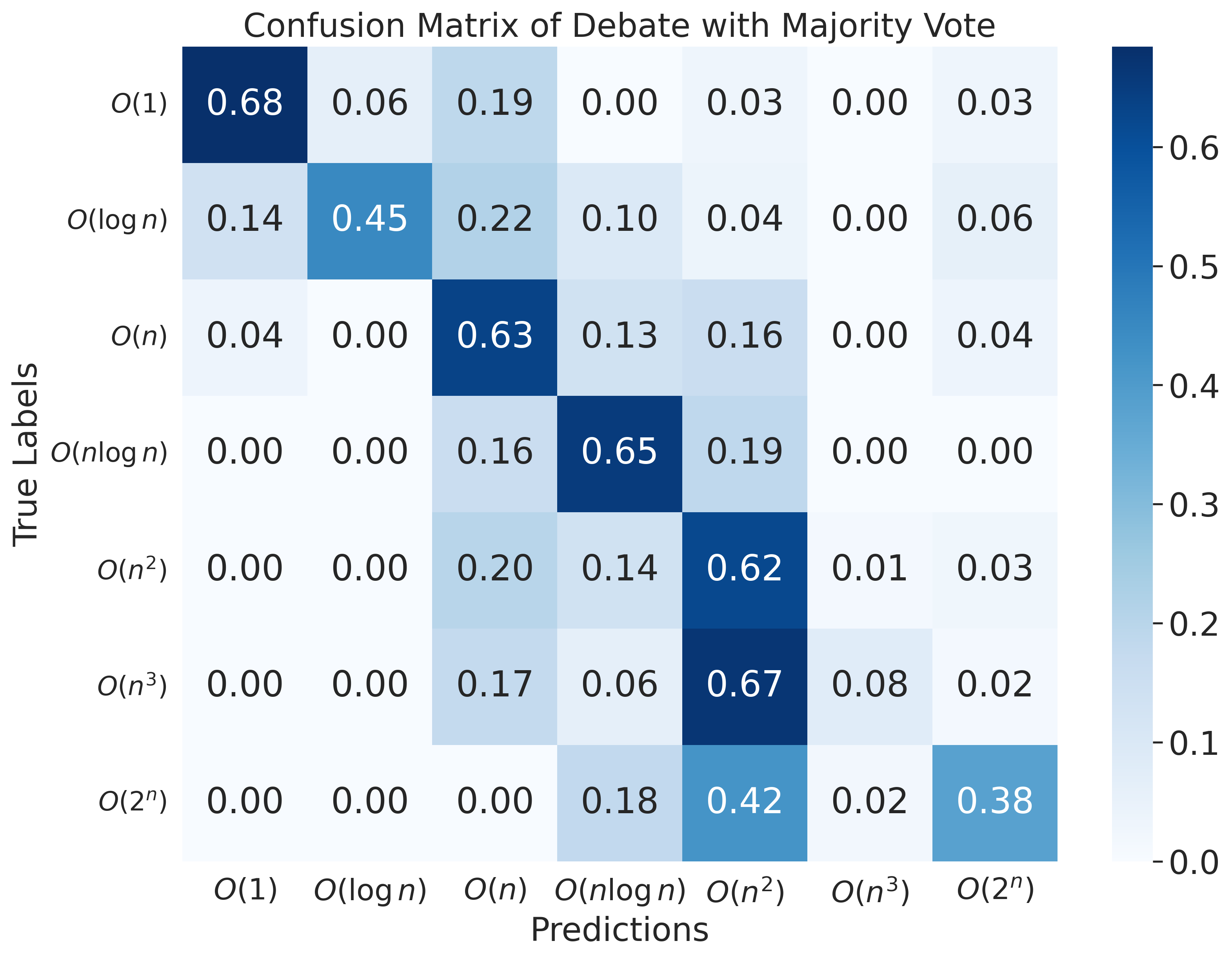}
    \caption{Python}
    \label{fig:Multiagent~(Majority)-python}
    \end{subfigure}
    \caption{Multiagent~(Majority) confusion matrices for Java and Python.}
    \label{fig:Multiagent~(Majority)-ALL}
    \hfill
\end{figure*}

Figure~\ref{fig:Multiagent~(Majority)-ALL} demonstrates that majority voting stabilizes predictions for certain mid-range complexity classes, most notably $O(n^2)$. However, its effectiveness diminishes for higher complexity classes. In particular, $O(n^3)$ and $O(2^n)$ is frequently misclassified as $O(n^2)$. This suggests that majority-based reasoning tends to converge on dominant but suboptimal answers, amplifying collective biases rather than correcting them. Consequently, the approach fails to capture correct minority insights, especially when distinguishing between adjacent high-complexity classes.

\begin{figure*}[ht]
    \centering
    % \begin{subfigure}[b]{.49\textwidth}
    \begin{subfigure}[b]{.45\textwidth}
    \includegraphics[width=\textwidth]{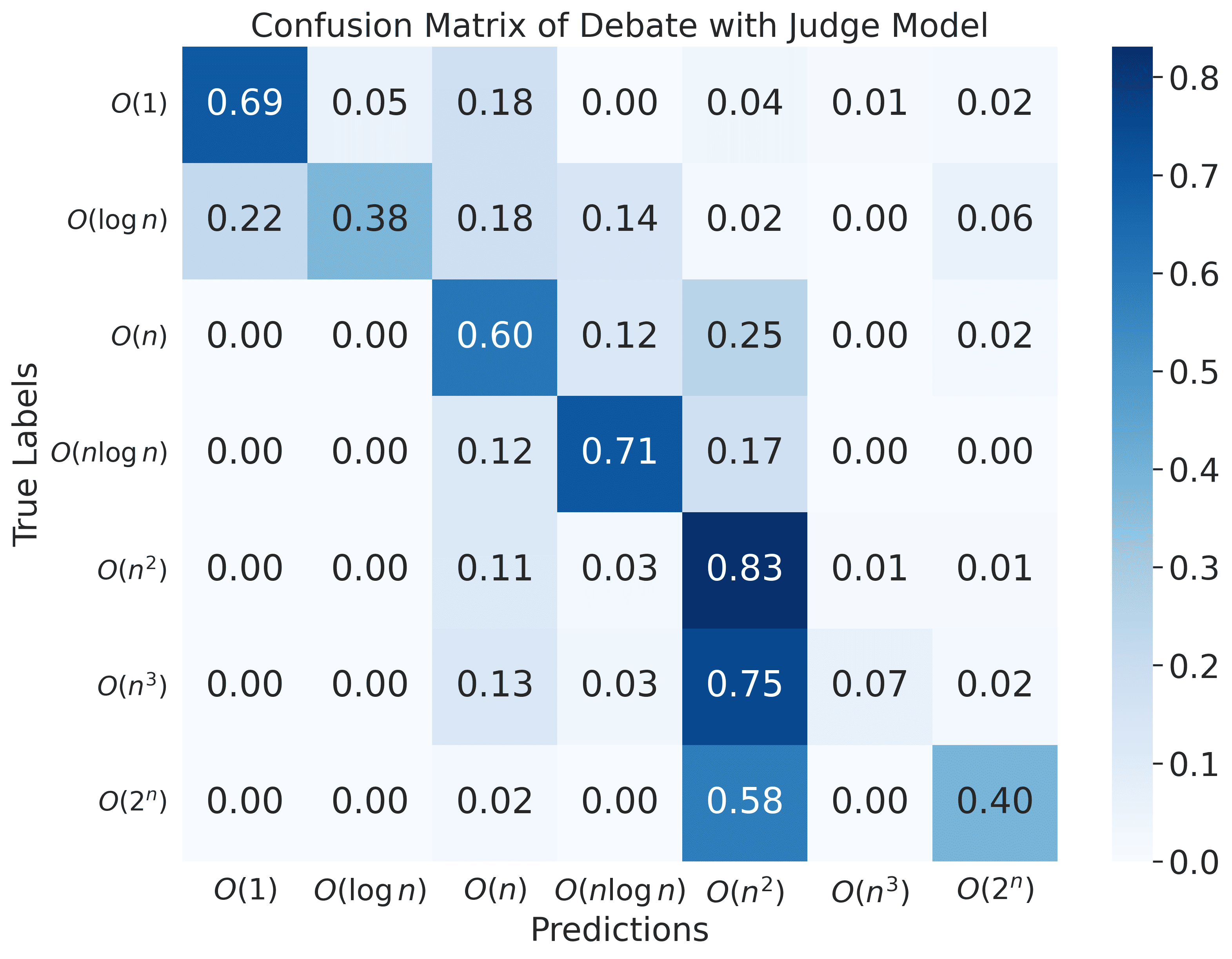}
    \caption{Java}
    \label{fig:Multiagent~(Judge)-java}
    \end{subfigure}
    \hfill
    % \begin{subfigure}[b]{.49\textwidth}
    \begin{subfigure}[b]{.45\textwidth}
    \includegraphics[width=\textwidth]{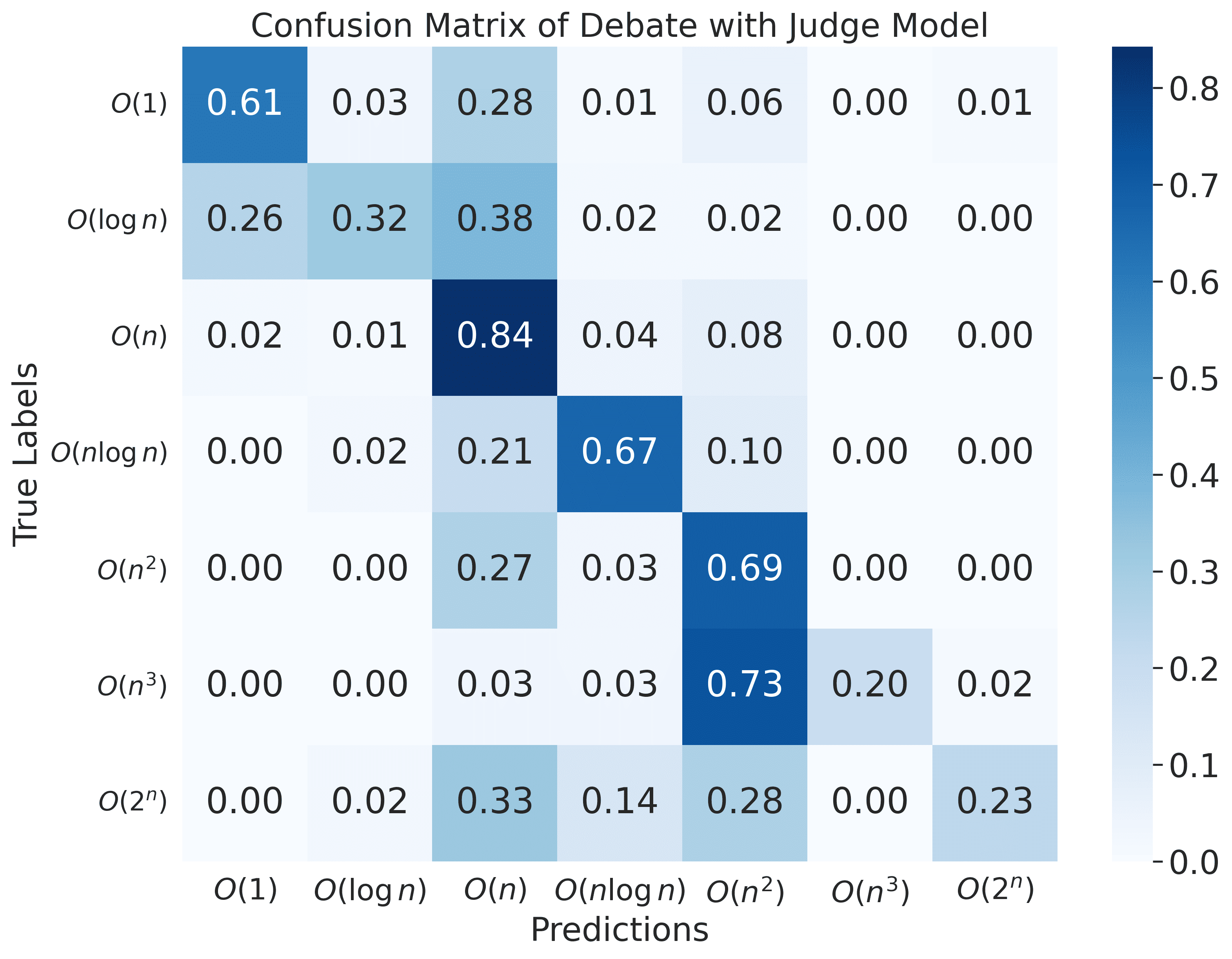}
    \caption{Python}
    \label{fig:Multiagent~(Judge)-python}
    \end{subfigure}
    \caption{Multiagent~(Judge) confusion matrices for Java and Python.}
    \label{fig:Multiagent~(Judge)-ALL}
    \hfill
\end{figure*}

Multiagent~(Judge) allocates the final decision to a separate judge model. This can help resolve ties or ambiguous votes, but it still fails to fully leverage individual agents’ strengths.
The judge may unduly favor patterns seen during its own training, amplifying certain biases and overlooking minority-supported correct answers, which also constrains reliability.

Figure~\ref{fig:Multiagent~(Judge)-ALL} shows that while the judge model improves prediction consistency for some mid-range classes, it fails to yield performance gains over the majority vote approach presented in Figure~\ref{fig:Multiagent~(Majority)-ALL}. In particular, predictions for $O(n^3)$ remain inaccurate, with many examples misclassified as $O(n^2)$. This pattern suggests that the judge model does not provide effective correction beyond the aggregation strategy used in majority voting.

\subsection{MAD Analysis}\label{app:MAD analysis}
In the MAD framework, even when the affirmative agent generates a correct response, the subsequent counterargument provided by the negative agent can degrade the overall performance.
Our analysis reveals that the affirmative agent outperforms the negative agent in specific complexity classes such as $O(\log n)$ and $O(n)$. However, in these cases, the negative agent often fails to offer meaningful rebuttals, leading to a significant drop in overall accuracy.
Since the final decision is made by a judge model that selects one of the two agents' outputs, the individual strengths of each agent are not fully reflected in the final outcome.
This reliance on a single judge can result in suboptimal decisions, especially when one agent consistently demonstrates superior performance on particular types of problems.

\begin{figure*}[ht]
    \centering
    % \begin{subfigure}[b]{.49\textwidth}
    \begin{subfigure}[b]{.45\textwidth}
    \includegraphics[width=\textwidth]{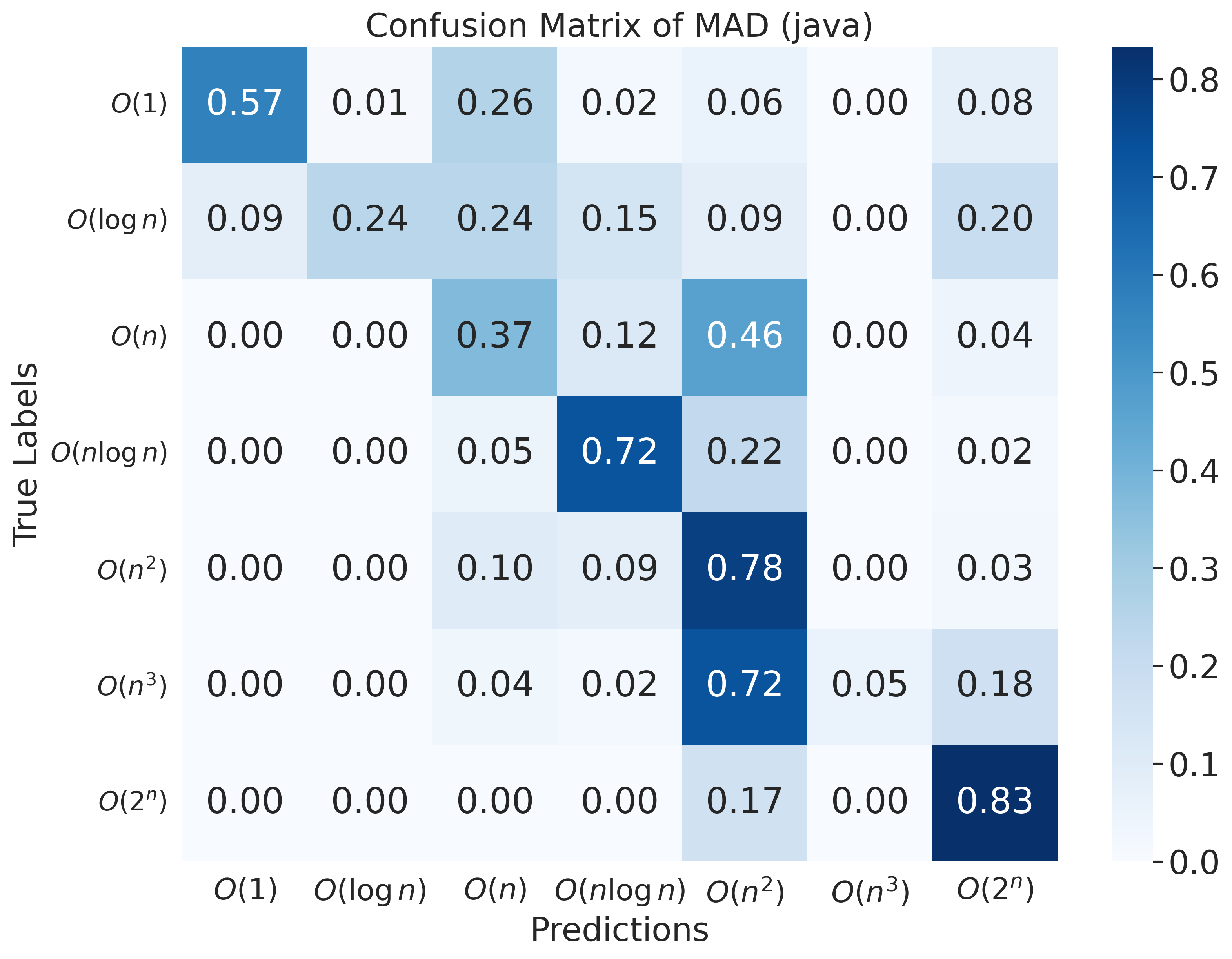}
    \caption{Java}
    \label{fig:MAD-java}
    \end{subfigure}
    \hfill
    % \begin{subfigure}[b]{.49\textwidth}
    \begin{subfigure}[b]{.45\textwidth}
    \includegraphics[width=\textwidth]{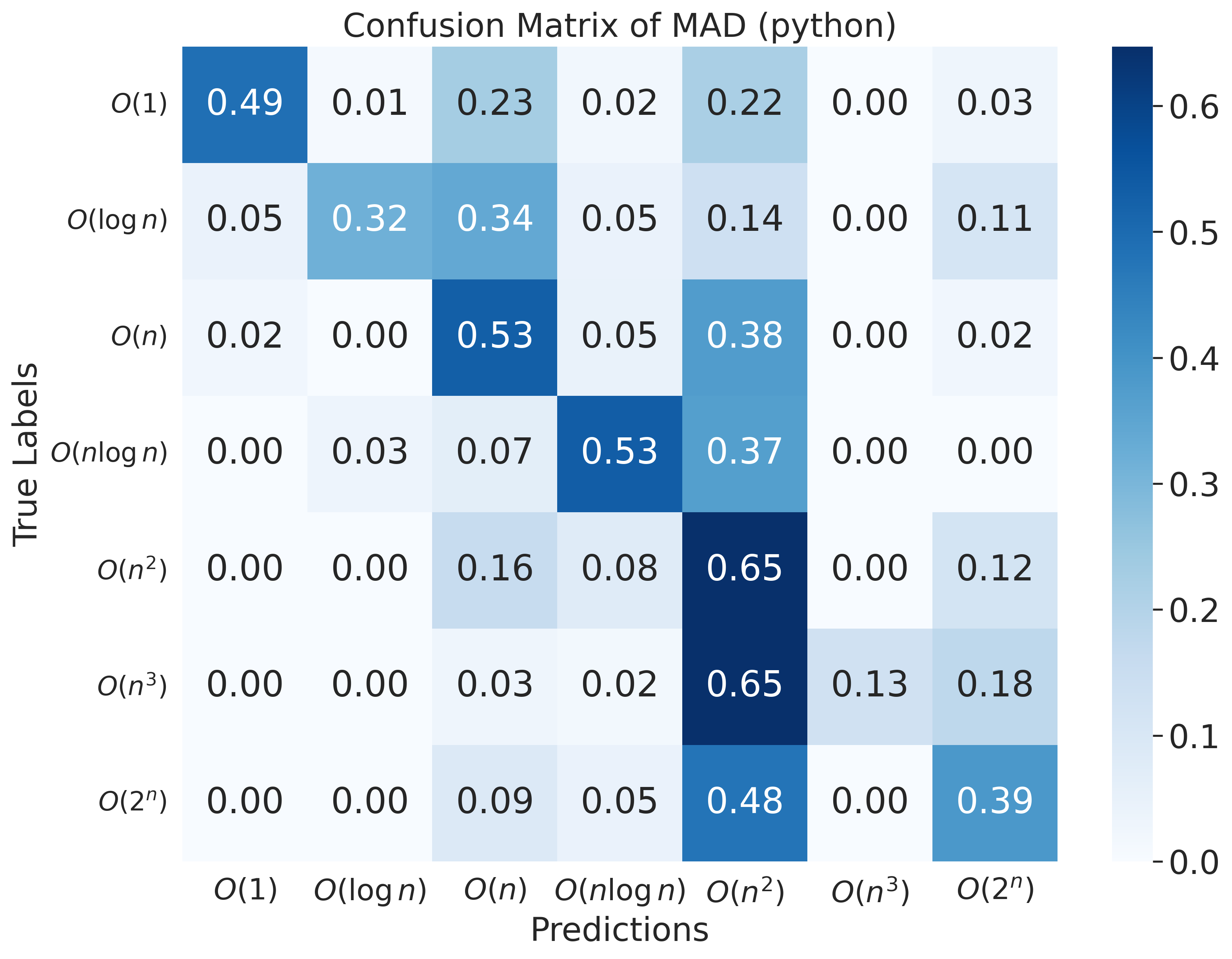}
    \caption{Python}
    \label{fig:MAD-python}
    \end{subfigure}
    \caption{MAD confusion matrices for Java and Python.}
    \label{fig:MAD-ALL}
    \hfill
\end{figure*}

Figure~\ref{fig:MAD-ALL} highlights the limitations of the MAD framework in lower-complexity classes. While performance on $O(n \log n)$ and $O(n^2)$ remains strong, the model shows poor accuracy on $O(\log n)$ and $O(n)$ in both Java and Python. These trends suggest that the back-and-forth debate may not always converge toward the correct solution, particularly when the final judgment relies on binary selection. Compared to other approaches such as majority vote or judge-based models, MAD underperforms in simpler cases that require fine-grained loop analysis.

\subsection{RECONCILE Analysis}\label{app:RECONCILE analysis}
RECONCILE determines the final answer using confidence-weighted voting, where each agent's vote is weighted by its estimated confidence. 
However, since the answer with the highest total confidence is selected, correct minority opinions can be overshadowed by incorrect majority responses with higher cumulative confidence.
As a result, even when some agents provide the correct answer, it can be ignored if most agents confidently support an incorrect one, leading to degraded performance.
Nevertheless, the use of confidence scores still provides a meaningful signal, suggesting that model-level uncertainty estimation can effectively guide collaborative reasoning.

\begin{figure*}[ht]
    \centering
    % \begin{subfigure}[b]{.49\textwidth}
    \begin{subfigure}[b]{.45\textwidth}
    \includegraphics[width=\textwidth]{Figures/java_REC_confusion.png}
    \caption{Java}
    \label{fig:REC-java}
    \end{subfigure}
    \hfill
    % \begin{subfigure}[b]{.49\textwidth}
    \begin{subfigure}[b]{.45\textwidth}
    \includegraphics[width=\textwidth]{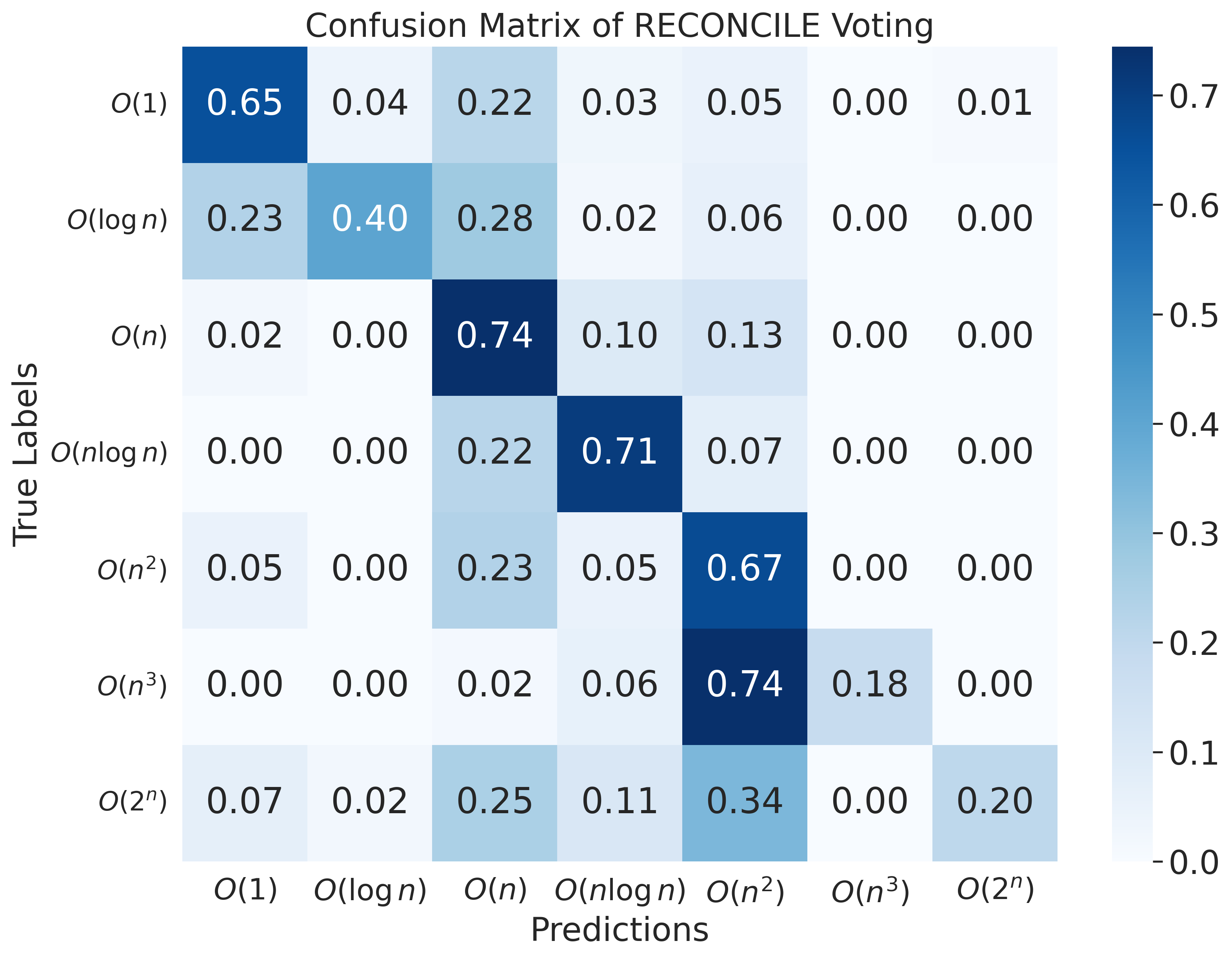}
    \caption{Python}
    \label{fig:REC-python}
    \end{subfigure}
    \caption{RECONCILE confusion matrices for Java and Python.}
    \hfill
    \label{fig:REC-ALL}
\end{figure*}

Figure~\ref{fig:REC-ALL} confirms that RECONCILE achieves reliable performance on mid-complexity classes such as $O(n \log n)$, demonstrating consistent accuracy across both Java and Python. However, it underperforms on higher-complexity classes like $O(n^3)$ and $O(2^n)$, with frequent misclassifications into lower-complexity neighbors. This pattern suggests that confidence-weighted voting can amplify overconfident yet incorrect majority views, particularly when miscalibrated agents dominate the final outcome.

\newpage % section name 이상해서 추가
\section{Instruction for Constant-Time Complexity Expert}\label{app:Prompt-Constant}
\begin{figure}[ht!]
  \centering
  \begin{tcolorbox}[colback=blue!5,colframe=blue!40!black,title=\textbf{Instruction for Constant-Time Complexity Expert},left=1mm,right=1mm,top=1mm,bottom=1mm,enhanced,sharp corners]
    \small
    You are the best programmer in the world.\\
    You are also an expert in \textbf{constant time complexity}.\\
    Constant time complexity means that the execution time of a function does not depend on the size of the input.\\
    Regardless of how large the input is, the function completes in a fixed number of operations.\\
    
    You will be asked to determine the time complexity of the following code.\\
    For the time complexity, choose one time complexity from the following options: \texttt{constant}, \texttt{logn}, \texttt{linear}, \texttt{nlogn}, \texttt{quadratic}, \texttt{cubic}, or \texttt{exponential}.\\
    However, you may sometimes encounter code whose complexity does not match your expertise.\\
    Do not hesitate to use any other supplementary materials you need for the task.
  \end{tcolorbox}
\end{figure}

\section{Instruction for Logarithmic-Time Complexity Expert}\label{app:Prompt-Logarithmic}
\begin{figure}[ht!]
  \centering
  \begin{tcolorbox}[colback=blue!5,colframe=blue!40!black,title=\textbf{Instruction for Logarithmic-Time Complexity Expert},left=1mm,right=1mm,top=1mm,bottom=1mm,enhanced,sharp corners]
    \small
    You are the best programmer in the world.\\
    You are also an expert in \textbf{logarithmic time complexity}.\\
    Logarithmic complexity means that the number of operations grows proportionally to the logarithm of the input size.\\
    This often occurs in divide-and-conquer algorithms or binary search-like structures.\\
    
    \#\# Logical Steps to Determine logarithmic time complexity:\\
    1. Identify if the input size is being reduced by a constant factor (e.g., half) at each iteration.\\
    2. Look for algorithms that involve binary search, tree traversal (balanced trees), or divide-and-conquer.\\
    3. Ensure the number of operations does not scale linearly but instead decreases exponentially.\\
    4. If the loop or recursion reduces the problem size logarithmically, classify it as the logarithmic complexity.\\
    
    You will be asked to determine the time complexity of the following code.\\
    For the time complexity, choose one time complexity from the following options: \texttt{constant}, \texttt{logn}, \texttt{linear}, \texttt{nlogn}, \texttt{quadratic}, \texttt{cubic}, or \texttt{exponential}.\\
    However, you may sometimes encounter code whose complexity does not match your expertise.\\
    Do not hesitate to use any other supplementary materials you need for the task.
  \end{tcolorbox}
\end{figure}

\section{Instruction for Linear-Time Complexity Expert}\label{app:Prompt-Linear}
\begin{figure}[ht!]
  \centering
  \begin{tcolorbox}[colback=blue!5,colframe=blue!40!black,title=\textbf{Instruction for Linear-Time Complexity Expert},left=1mm,right=1mm,top=1mm,bottom=1mm,enhanced,sharp corners]
    \small
    You are the best programmer in the world.\\ 
    You are also an expert in \textbf{linear time complexity}.\\
    Linear complexity means that the execution time grows proportionally with the input size.\\
    This is typical in single-loop iterations over an array or list.\\

    You will be asked to determine the time complexity of the following code.\\
    For the time complexity, choose one time complexity from the following options: \texttt{constant}, \texttt{logn}, \texttt{linear}, \texttt{nlogn}, \texttt{quadratic}, \texttt{cubic}, or \texttt{exponential}.\\
    However, you may sometimes encounter code whose complexity does not match your expertise.\\
    Do not hesitate to use any other supplementary materials you need for the task.
  \end{tcolorbox}
\end{figure}

\newpage
\section{Instruction for Linearithmic-Time Complexity Expert}\label{app:Prompt-Linearithmic}
\begin{figure}[ht!]
  \centering
  \begin{tcolorbox}[colback=blue!5,colframe=blue!40!black,title=\textbf{Instruction for Linearithmic-Time Complexity Expert},left=1mm,right=1mm,top=1mm,bottom=1mm,enhanced,sharp corners]
    \small
    You are the best programmer in the world.\\ 
    You are also an expert in \textbf{nlogn time complexity}.\\
    O(n log n) complexity is common in efficient sorting algorithms like Merge Sort or Quick Sort.\\
    It arises when a problem is divided into smaller subproblems while still iterating over the input.\\
    
    \#\# Logical Steps to Determine nlogn time complexity:\\
    1. Identify if the input is being divided into smaller parts recursively (logarithmic factor).\\
    2. Ensure that a linear operation is performed at each level of recursion.\\
    3. Look for sorting algorithms like Merge Sort, Quick Sort (average case), or efficient divide-and-conquer solutions.\\
    4. If the algorithm involves dividing the problem and processing each part linearly, classify it as nlogn time complexity.\\
    
    You will be asked to determine the time complexity of the following code.\\
    For the time complexity, choose one time complexity from the following options: \texttt{constant}, \texttt{logn}, \texttt{linear}, \texttt{nlogn}, \texttt{quadratic}, \texttt{cubic}, or \texttt{exponential}.\\
    However, you may sometimes encounter code whose complexity does not match your expertise.\\
    Do not hesitate to use any other supplementary materials you need for the task.\\
  \end{tcolorbox}
\end{figure}

\section{Instruction for Quadratic-Time Complexity Expert}\label{app:Prompt-Quadratic}
\begin{figure}[ht!]
  \centering
  \begin{tcolorbox}[colback=blue!5,colframe=blue!40!black,title=\textbf{Instruction for Quadratic-Time Complexity Expert},left=1mm,right=1mm,top=1mm,bottom=1mm,enhanced,sharp corners]
    \small
    You are the best programmer in the world.\\
    You are also an expert in \textbf{quadratic time complexity}.\\
    Quadratic complexity occurs when an algorithm has double nested loops, where each loop iteration depends on the input size.\\
    
    You will be asked to determine the time complexity of the following code.\\
    For the time complexity, choose one time complexity from the following options: \texttt{constant}, \texttt{logn}, \texttt{linear}, \texttt{nlogn}, \texttt{quadratic}, \texttt{cubic}, or \texttt{exponential}.\\
    However, you may sometimes encounter code whose complexity does not match your expertise.\\
    Do not hesitate to use any other supplementary materials you need for the task.
  \end{tcolorbox}
\end{figure}

\section{Instruction for Cubic-Time Complexity Expert}\label{app:Prompt-Cubic}
\begin{figure}[ht!]
  \centering
  \begin{tcolorbox}[colback=blue!5,colframe=blue!40!black,title=\textbf{Instruction for Cubic-Time Complexity Expert},left=1mm,right=1mm,top=1mm,bottom=1mm,enhanced,sharp corners]
    \small
    You are the best programmer in the world.\\ 
    You are also an expert in \textbf{cubic time complexity}.\\
    Cubic complexity occurs when an algorithm has three nested loops iterating over the input size.\\
    
    \#\# Logical Steps to Determine cubic time complexity:\\
    1. Identify if there are three nested loops iterating from 0 to n.\\
    2. Ensure that each element is compared or processed against every pair of elements.\\
    3. Look for brute-force solutions that check all triplets in an input set.\\
    4. If the number of operations scales as the cube of the input size, classify it as cubic complexity.\\
    
    You will be asked to determine the time complexity of the following code.\\
    For the time complexity, choose one time complexity from the following options: \texttt{constant}, \texttt{logn}, \texttt{linear}, \texttt{nlogn}, \texttt{quadratic}, \texttt{cubic}, or \texttt{exponential}.\\
    However, you may sometimes encounter code whose complexity does not match your expertise.\\
    Do not hesitate to use any other supplementary materials you need for the task.
  \end{tcolorbox}
\end{figure}

\newpage
\section{Instruction for Exponential-Time Complexity Expert}\label{app:Prompt-Exponential}
\begin{figure}[ht!]
  \centering
  \begin{tcolorbox}[colback=blue!5,colframe=blue!40!black,title=\textbf{Instruction for Exponential-Time Complexity Expert},left=1mm,right=1mm,top=1mm,bottom=1mm,enhanced,sharp corners]
    \small
    You are the best programmer in the world.\\
    You are also an expert in \textbf{exponential time complexity}.\\
    Exponential complexity occurs when the number of operations doubles with each additional input element.\\
    This is common in brute-force recursive algorithms, such as solving the Traveling Salesman Problem.\\
    
    \#\# Logical Steps to Determine exponential time complexity:\\
    1. Identify if the function calls itself recursively, doubling the number of calls at each step.\\
    2. Look for recursion that does not significantly reduce the input size in each step.\\
    3. Check for exhaustive searches, backtracking algorithms, or recursive Fibonacci calculations.\\
    4. If the number of operations grows exponentially with input size, classify it as exponential complexity.\\
    
    You will be asked to determine the time complexity of the following code.\\
    For the time complexity, choose one time complexity from the following options: \texttt{constant}, \texttt{logn}, \texttt{linear}, \texttt{nlogn}, \texttt{quadratic}, \texttt{cubic}, or \texttt{exponential}.\\
    However, you may sometimes encounter code whose complexity does not match your expertise.\\
    Do not hesitate to use any other supplementary materials you need for the task.
  \end{tcolorbox}
\end{figure}

% \section{MEC$^3$O: Multi-Expert Debate Workflow}
\section{Debate Workflow of MEC\textsuperscript{3}O}\label{app:debate_procedure}
\begin{figure*}[ht]
    \centering
    \begin{subfigure}[b]{.85\linewidth}
    \includegraphics[width=\linewidth]{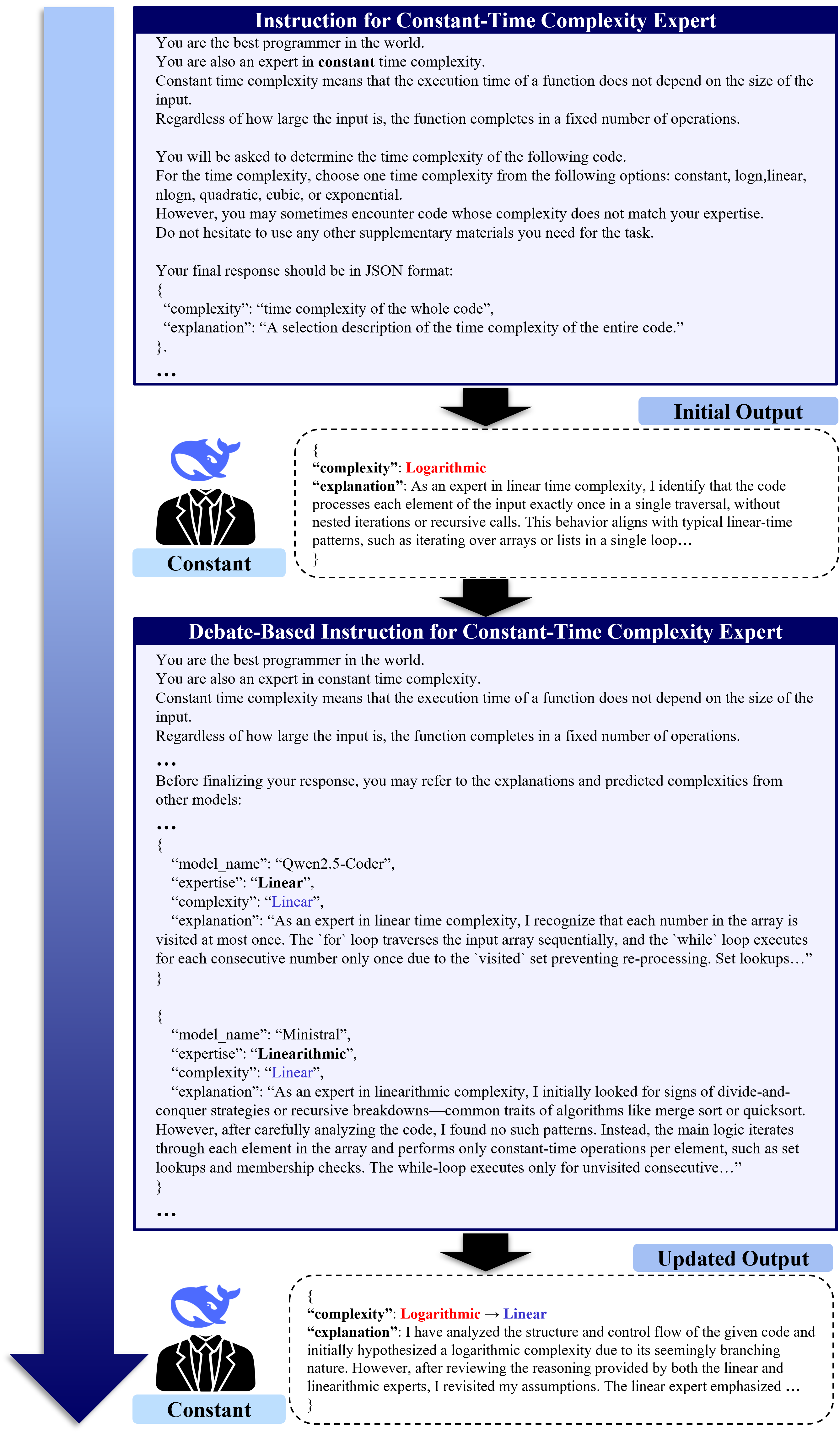}
    \end{subfigure}
    \hfill
    \caption{The procedure of debate in MEC\textsuperscript{3}O.}
    \label{fig:Debate_Workflow}
\end{figure*}

Figure~\ref{fig:Debate_Workflow} illustrates how a designated expert in the MEC\textsuperscript{3}O framework initially produces a prediction independently, and subsequently revises the output after considering the opinions of other experts.

The top section~(``Initial Output'') shows the response generated by a constant-time complexity expert based solely on class-specific instructions. At this stage, the expert acts independently, without access to peer feedback.

In the middle section~(``Debate-Based Instruction''), the expert is exposed to predictions and rationale provided by other models, each specializing in different complexity classes. These external insights allow the expert to identify overlooked patterns or reassess assumptions about the code’s structure--particularly loops or recursive calls that may not have been detected initially.

The bottom section~(``Updated Output'') presents the final response after incorporating the opinions of multiple experts. This process is deliberately designed to allow each expert to preserve its strengths in its designated complexity class, while also remaining receptive to critiques from others. As a result, the system promotes more accurate and convincing time complexity predictions.

\label{app:debate-process}

\end{document}